\documentclass[manuscript]{acmart}

\AtBeginDocument{%
  \providecommand\BibTeX{{%
    \normalfont B\kern-0.5em{\scshape i\kern-0.25em b}\kern-0.8em\TeX}}}

\setcopyright{acmcopyright}
\copyrightyear{2022}
\acmYear{2022}
\acmDOI{10.1145/1122445.1122456}

\usepackage{multirow}
\usepackage{mathtools}
\usepackage{xspace}
\usepackage{subfig}
\usepackage{float}
\usepackage{placeins}
\usepackage{hhline}

\acmConference[ACM FAccT 2022]{ACM Conference on Fairness, Accountability, and Transparency 2022}{21-24 June, 2022}{Seoul, South Korea}
\acmBooktitle{ACM Conference on Fairness, Accountability, and Transparency 2022}

\newcommand{\carbondioxide}{CO\textsubscript{2}\xspace}
\newcommand{\carboneq}{CO\textsubscript{2}eq}

\copyrightyear{2022}
\acmYear{2022}
\setcopyright{rightsretained}
\acmConference[FAccT '22]{2022 ACM Conference on Fairness, Accountability, and Transparency}{June 21--24, 2022}{Seoul, Republic of Korea}
\acmBooktitle{2022 ACM Conference on Fairness, Accountability, and Transparency (FAccT '22), June 21--24, 2022, Seoul, Republic of Korea}\acmDOI{10.1145/3531146.3533234}
\acmISBN{978-1-4503-9352-2/22/06}

\begin{document}


\title{Measuring the Carbon Intensity of AI in Cloud Instances}

\author{Jesse Dodge}
\affiliation{%
  \institution{Allen Institute for AI}
  \country{USA}
  }
  
\author{Taylor Prewitt}
\affiliation{%
  \institution{University of Washington}
  \country{USA}
  }
  \email{prewitt425@gmail.com}

\author{Remi Tachet Des Combes}
\affiliation{%
  \institution{Microsoft Research Montreal}
  \country{USA}
  }
  \email{Remi.Tachet@microsoft.com}

\author{Erika Odmark}
\affiliation{%
  \institution{Microsoft}
  \country{USA}
  }
  \email{Erika.Odmark@microsoft.com}

\author{Roy Schwartz}
\affiliation{%
  \institution{Hebrew University of Jerusalem}
  \country{Israel}
}
\email{roy.schwartz1@mail.huji.ac.il}

\author{Emma Strubell}
\affiliation{%
  \institution{Carnegie Mellon University}
  \country{USA}
}
\email{strubell@cmu.edu}

\author{Alexandra Sasha Luccioni}
\affiliation{%
  \institution{Hugging Face}
  \country{USA}
  }
  \email{sasha.luccioni@huggingface.co}

\author{Noah A. Smith}
\affiliation{%
  \institution{Allen Institute for AI and University of Washington}
  \country{USA}
  }
  \email{noah@allenai.org}

\author{Nicole DeCario}
\affiliation{%
  \institution{Allen Institute for AI}
  \country{USA}
  }
  \email{nicoled@allenai.org}

\author{Will Buchanan}
\affiliation{%
  \institution{Microsoft}
  \country{USA}
  }
  \email{wibuchan@microsoft.com}

\renewcommand{\shortauthors}{Dodge et al.}

\begin{abstract}
The advent of cloud computing has provided people around the world with unprecedented access to computational power and enabled rapid growth in technologies such as machine learning,
the computational demands of which incur a high energy cost and a commensurate carbon footprint. As a result, recent scholarship has called for better estimates of the greenhouse gas impact of AI: data scientists today do not have easy or reliable access to measurements of this information, which precludes development of actionable tactics. We argue that cloud providers presenting information about software carbon intensity to users is a fundamental stepping stone towards minimizing emissions.

In this paper, we provide a framework for measuring software carbon intensity, and propose to measure operational carbon emissions by using location-based and time-specific marginal emissions data per energy unit. We provide measurements of operational software carbon intensity for a set of modern models covering natural language processing and computer vision applications, and a wide range of model sizes, including pretraining of a 6.1 billion parameter language model.
We then evaluate a suite of approaches for reducing emissions on the Microsoft Azure cloud compute platform: using cloud instances in different geographic regions, using cloud instances at different times of day, and dynamically pausing cloud instances when the marginal carbon intensity is above a certain threshold. We confirm previous results that the geographic region of the data center plays a significant role in the carbon intensity for a given cloud instance, and find that choosing an appropriate region can have the largest operational emissions reduction impact. We also present new results showing that the time of day has meaningful impact on operational software carbon intensity.Finally, we conclude with recommendations for how machine learning practitioners can use software carbon intensity information to reduce environmental impact.




\end{abstract}



\maketitle

\section{Introduction}

Climate change is an increasing threat to life on our planet, which disproportionately impacts the most disadvantaged communities and fragile ecosystems~\cite{masson2018global}. One of the main drivers of climate change is carbon dioxide, or \carbondioxide, which contributes to the greenhouse effect by trapping the heat from the sun within the atmosphere without letting it dissipate. \carbondioxide  (and other types of \textit{greenhouse gases}, such as methane and ozone) are emitted by many sources, some natural but most man-made, such as the burning of oil and gas for transportation and heating or for industrial processes such as smelting. In 2018, it was estimated that global data center energy use represented close to 1\% of global energy usage~\cite{masanet2020recalibrating}. While it is not yet known what proportion of data center use is for training artificial intelligence (AI) models, it is undeniable that AI and its sub-fields have grown dramatically in recent years, with no sign of slowing down~\cite{schwartz2020green,thompson2020computational}. 
While a number of papers have addressed the \carbondioxide emissions produced by AI (e.g., ~\cite{lacoste2019quantifying, ligozat2021unraveling, patterson2021carbon,patterson2022carbon}), the extent and provenance of \carbondioxide emissions in the field is still under-explored. Nonetheless, a common theme of previous work is that it aims to estimate the emissions produced by training AI models, or carrying out the accompanying neural architecture search (NAS) process, based on coarse measures such as \carbondioxide emissions of electricity used in the region where the computations were carried out (e.g., \cite{strubell2019energy}), or post-hoc analyses using information that is not publicly available (e.g., ~\cite{patterson2021carbon}). 

With an increasing amount of AI model training being done on cloud compute instances, 
reducing the emissions generated by these workloads will be key to reducing our carbon footprint as a field. However, to reduce greenhouse gas emissions from cloud computing, we need consider the role of two types of actors: the cloud provider (such as Microsoft Azure, Google's GCP, or Amazon's AWS) and the user who reserves and uses cloud resources (e.g., an AI researcher training a model on a cloud instance, or a company hosting a website). Typically, the provider's motivation is to build a system where users can access the computing power and storage that best meets their needs.
The user, on the other hand, is motivated by some end task which requires computing power, such as running a set of experiments or putting a model into production. Often the user will first consider the minimal computational requirements to achieve their goals, then later ease-of-use features relating to transfer speed or extra storage depending on available budget.
Driven by these motivations, providers and users can each take actions to meet their goals: providers can build data centers and set up APIs to enable users' access and accounting, while users can choose their cloud provider, which region to use, and the number and type of cloud instances required for their end task at a given point in time. Based on these stakeholders and motivations, in this work we address the following research questions: 1) how should we measure and report operational carbon costs of AI workloads? And 2) can we shift computation spatially and temporally to mitigate emissions?

In this article, we introduce the first tool to estimate the real-time \carbondioxide emissions impact of instances on a cloud computing 
platform.
The tool calculates operational carbon emissions by using location-based and time-specific marginal emissions data per energy unit. Using the tool, we explore several case studies on the Microsoft Azure cloud compute platform spanning the areas of natural language processing (NLP) and computer vision, estimating the carbon intensity of training a variety of commonly used machine learning models. We also explore two avenues for users of cloud instances to reduce their \carbondioxide using this tool by: (1) Changing the region of compute and (2) changing the time of day during which the model is run. While the former has been recognized by prior work \citep{lacoste2019quantifying,google-region}, we are the first to address the latter to the best of our knowledge. Further, our tool makes it possible to automatically schedule jobs in order to reduce their carbon footprint by leveraging these differences in carbon intensity due to time and geographic location. Finally, we provide guidance regarding what should be measured and how, following the Green Software Foundation's guidelines regarding Software Carbon Intensity (SCI), and suggest future areas of research to improve the state of carbon estimation and reporting in AI.

\section{Related work} \label{sec:related-work}
Attention was first drawn to the environmental impact of AI research by the seminal work of Strubell et al.~\cite{strubell2019energy}, which quantified the emissions produced by training a Transformer model with Neural Architecture search, finding it to be comparable to the lifetime carbon emissions of five cars. \citet{patterson2021carbon} presented some updated analyses of similar experiments, including popular architectures like T5~\cite{raffel2019exploring} and BERT~\cite{devlin2019bert}, analyzing \carbondioxide emissions as a factor of their energy consumption, carbon intensity of training servers, etc. Other work such as Green AI~\cite{schwartz2020green} delved further into inequality of access to computational resources within the research community, and advocated for the inclusion of efficiency evaluation alongside accuracy as a primary evaluation criterion. Much existing and ongoing work on quantifying the environmental footprint of ML has been focused on estimating the \carbondioxide emissions of model training. This is a more straightforward endeavor compared to other stages both upstream and downstream from the training process, given that it is well-defined in time and its emissions can be measured in real-time with tools like Code Carbon~\cite{schmidt2021codecarbon} and Carbon Tracker~\cite{anthony2020carbontracker} or estimated post-hoc using tools such as ML \carbondioxide Impact Tracker~\cite{lacoste2019quantifying}.
Our tool builds upon this work by making carbon tracking on cloud instances possible, enabling a larger portion of ML model training work to profit from fine-grained carbon estimation.  However, recent work has found that their results vary significantly and are not fully representative of the true emissions incurred by training~\cite{bannour2021evaluating}. Perhaps most similar to our work, EnergyVis~\cite{shaikh21energyviz} is an interactive tool for visualizing and comparing energy consumption of ML models as a function of hardware and physical location (U.S. state), given metadata about a model's energy use per epoch. Other studies have gone beyond simply tracking the emissions from training models, aiming to quantify the emissions resulting from manufacturing computing hardware~\cite{gupta2021chasing}, the broader impacts of sustainable AI~\cite{wu2021sustainable}, and the methodologies used to assess those impacts~\cite{ligozat2021unraveling, kaack2021aligning}. Building upon this research, efforts have also been made to certify systems as being socially- and environmentally-conscious~\cite{gupta2020secure}, working towards comparing both the environmental costs and potential benefits of AI models in order to paint a more holistic picture of AI.

Major technology companies have also been increasingly committed to reducing their emissions, largely via the purchase of Renewable Energy Credits (RECs), which involves directly buying quantities of energy produced by renewable sources, translating into carbon reductions under the assumption that the clean energy is displacing an equivalent amount of electricity produced by non-renewable methods~\cite{gillenwater2008redefining}. Many cloud providers, from Google Cloud Platform to Microsoft Azure, therefore claim that they are now ``carbon-neutral,'' given that they offset the entirety of the emissions of their cloud centers, though we must be wary of the precise provenance of RECs, and the details of how each organization defines ``zero'' net emissions \cite{nature-climate-pledges-21}. This is complemented by efforts to mitigate the actual \carbondioxide emissions of the compute regions themselves, with several server locations partially powered by renewable energy sources such as solar and wind~\cite{azure-sustain, aws-sustain, google-sustain} and giving users the necessary tools to pick compute regions with a smaller carbon footprint~\cite{azure-dashboard,google-region}, which are often tied to the amount of low-carbon energy that is being purchased, and not the grid emissions intensity. It is important to note that the decision on when and where to deploy a workload should be based on a grid emissions signal, not the amount of emissions offset through market-based measures (e.g., green power purchase agreements (PPAs), renewable energy certificates (RECs), or other carbon offset mechanisms): purchasing clean energy is not the same as consuming clean energy.

\section{Reporting AI Carbon Intensity}
Carbon accounting and reporting is becoming increasingly common in ML, with conferences such as NeurIPS requesting that submissions report their emissions~\cite{neurips-checklist} and recent work reporting the emissions incurred~\cite{thoppilan2022lamda,sanh2021multitask}. However, it has yet to become the norm in our field, and we are still lacking systematic information regarding the environmental footprint of training ML models and how we can reduce it. In this paper, we argue that if members of the ML community had access to information about the \carbondioxide emissions of their actions, they could adapt their decisions to reduce these emissions while still meeting the computational needs for their end tasks.
In addition, providers building tools that enable users to track their \carbondioxide emissions directly aligns with providers' goals, as it will inform users' decisions without being overly burdensome. Any cloud provider that discloses this information to users will, in fact, be improving those customers' experiences, and likely increase usage of the platform. More specifically, we propose that, for a cloud user who wants to estimate their carbon footprint, the most salient information providers can report is the \carbondioxide emissions generated by their cloud instances.
Arguably the single most important contribution of this paper is the simplest: a presentation of the software carbon intensity (SCI) as a proxy for carbon emissions for a given cloud instance as it is running.

\subsection{Methodology: Computing \carbondioxide Intensity}\label{sec:methodology}
In this section we describe a method for estimating carbon intensity for cloud instances.
At a high level, this involves tracking electricity consumption of hardware related to a single cloud instance, and mapping that electricity usage to \carbondioxide emissions by using a grid-based carbon intensity.

As developed by the \href{https://github.com/Green-Software-Foundation/software_carbon_intensity/blob/dev/Software_Carbon_Intensity/Software_Carbon_Intensity_Specification.md}{Green Software Foundation}, the Software Carbon Intensity ($SCI$) is a rate, carbon emissions per one functional unit, or R. The equation used to calculate the $SCI$ value of a software system is therefore:
\begin{align}
SCI = ((E * I) + M) \mathrm{\ per\ } R
\end{align}
where: 
\begin{itemize}
    \item $E =$ Energy consumed by a software system. Specifically, we focus on energy consumption of Graphical Processing Units, or GPUs. 
    The units used are kilowatt-hours (kWh). 
    \item $I =$ Location-based marginal carbon emissions for the grid that powers the datacenter. \href{https://www.watttime.org/}{WattTime} provides measurements of grams of carbon dioxide equivalent per kilowatt-hour of electricity (g\carboneq/kWh)  
    \item $M =$ Embodied carbon (also referred to as ``embedded carbon'') is the amount of carbon emitted during the creation, usage, and disposal of a hardware device. When software runs on a device, a fraction of the total embodied emissions of the device is allocated to the software. 
    \item $R =$ Functional unit. In this instance, we are defining the functional unit as one machine learning training job, but it is extensible to other scenarios. 
\end{itemize}
The equation can be further refined to:
\begin{align}
SCI = (O + M) \mathrm{\ per\ } R
\end{align}
where $O = E * I$ calculates the operational emissions based on energy consumption ($E$) multiplied by the location-based and time-specific carbon intensity measurement ($I$). Once more this can be further refined to simply:
\begin{align}
SCI = C \mathrm{\ per\ } R
\end{align}
where $C = O + M$ is the software carbon intensity for a given cloud instance. In this paper, we focus on measuring operational emissions $O$, and leave measurement and accounting for embodied emissions due to specialized ML hardware such as GPUs to future work (see \S\ref{sec:future}).

The objective of the Green Software Foundation's Software Carbon Intensity (SCI) specification is to calculate and reduce a SCI score, based on carbon emissions reductions, rather than the currently-used market-based neutralization. Specifically, the SCI uses a "consequential" carbon accounting approach, which aims to quantify the marginal change in emissions caused by decisions or interventions. This differs from the commonly used "attributional" carbon accounting approach, which uses average carbon intensity data, meaning it does not provide the most actionable information to help reduce carbon emissions. Due to the myriad potential pitfalls of relying on market-based measures in place of actual reduction in emissions \cite{nature-climate-pledges-21}, it is not possible to reduce the SCI through carbon neutralization or carbon offsets. We assert that cloud providers should provide the SCI to developers and data scientists to help them make choices that reduce the carbon footprint of their ML workloads. 

\subsection{The Scope of our Tool: GPU Computation of a Single Cloud Instance}

Data centers typically comprise many computer systems and hardware components, including storage, GPUs, CPUs, and networking components.
We can break down the electricity usage for data centers into: 1) electricity that is used for  a single cloud instance, and 2) electricity that is used for the benefit of the whole data center. In this work we focus on the former, a single cloud instance; because of this, a reader should understand that our estimates of the electricity consumption and emissions are underestimates.\footnote{There is related work on estimating and reducing the electricity of data centers in general, e.g., \cite{gao2014machine,lazic2018data}.}

\paragraph{Electricity Consumption from a Single Cloud Instance}
The most accurate and popular AI models today are typically (deep) neural networks, which are most performant on specialized, highly parallelized, and often energy-intensive hardware~\cite{thompson2020computational}. 
The most common scenario is for AI workloads to run on graphics processing units (GPUs), which provide significant acceleration compared to CPUs (central processing units) but are more power-hungry (often consuming 250W-350W, compared to CPU consumption of 10-150W). Due to specialization to the matrix multiply operations at the core of neural network computations and a high rate of parallelization, GPUs can perform many more of these types of computations in the same amount of time as a CPU, but this increased computation throughput comes at an increased energy cost.
Thus in ML applications based on deep learning, the majority of the electricity consumption is due to the  GPU~\cite{pc-components,torbet-pc-energy}. While this result is fairly uncontroversial, we ran an experiment to confirm it. To do so, we trained a BERT-base model~\cite{devlin2019bert} on a single NVIDIA TITAN X GPU (12 GB) in a commodity server with two Intel Xeon E5-2630 v3 CPUs (2.4GHz) and 256GB RAM (16x16GB DIMMs) to measure the relative electricity consumption of different components. We trained the model using the original \href{https://github.com/google-research/bert}{code} provided by~\citet{devlin2019bert} on the language model pre-training task for 12 hours on one GPU, sampling the instantaneous energy use of the GPU, CPU and DRAM for each socket throughout that period, then averaging to get average power draw per component in watts. GPU energy draw was measured using \href{https://developer.nvidia.com/nvidia-system-management-interface}{\texttt{nvidia-smi}} and CPU and DRAM power draw were obtained using Intel's \href{https://web.archive.org/web/20190116164417/https:/01.org/rapl-power-meter}{RAPL}. Our measurements, in watts, are presented in Table~\ref{tab:gpu_vs_other}. As expected the GPU accounts for almost 3/4 of electricity consumption. 

\begin{table}[h]
\centering
\begin{tabular}{ |c|c|c|c|c|c|c| } 
 \hline
 \textit{Hardwa.} & GPU & CPU$_0$ & CPU$_1$ & DRAM$_0$ & DRAM$_1$ & \textit{Total}\\
 \hline
 \textit{Watts} & 187.1 & 22.9 & 9.3 & 23.0 & 9.3 & 251.6\\ 
 \hline
 \textit{Fraction} & 74\% & 9\% & 4\% & 9\% & 4\% & 100\% \\ 
 \hline
\end{tabular}
\caption{The electricity consumption, in watts and percentages, when training BERT base on a single NVIDIA TITAN X GPU (12GB), in a commodity server with two Intel Xeon E5-2630 v3 CPUs (2.4GHz) and 256GB RAM (16x16GB DIMMs). Power consumption is averaged across instantaneous measurements over 12 hours of training on using the masked language modeling objective. The GPU alone accounts for 74\% of the total energy consumption due to these components.}
\label{tab:gpu_vs_other}
\end{table}

\paragraph{Focus on GPUs}
In cloud datacenters, the CPUs, RAM, storage, and motherboards are often shared across multiple instances; while this provides the flexibility that makes the cloud so useful, it leads to technical limitations that make it difficult (and in some cases impossible) to properly estimate electricity consumption from these sources for a single instance.
However, GPUs are typically not shared across instances, and in fact for large AI workloads it's often the case that multiple GPUs are attached to a single instance, leading to an even greater proportion of the total energy consumption being used by the GPUs.
Thus, it is relatively easy to measure the GPU electricity consumption for a single instance, while it is not for other components.
For this reason, and because they typically consume the majority of electricity in AI workloads, in this work we only measure GPU electricity consumption.
We recognize this is a first step towards a more complete measurement, and provide further discussion in the next section.\footnote{We note that our conclusions drawn from experiments and analyses on time-shifting and location-shifting are still applicable with tools that measure more electricity than just the GPU.}

\paragraph{Other sources of \carbondioxide}
Data centers have a number of electricity uses that are important, but will not be covered by our tool.
According to the U.S. Department of Energy: ``The electricity consumed in these data centers is mainly by the equipment (50\%) and HVAC (25\%–40\%)'' \cite{udept-energy}. Such other sources of emissions can be accounted for using methods such as Power Usage Effectiveness (PUE), which can be used to describe the proportion of electricity consumption by the computing equipment vs. other sources. For a given datacenter, this can be turned into a factor which can be multiplied against the electricity consumption of computing equipment to get an estimate of the total consumption.
Some companies have highlighted particularly low PUEs, such as Google claiming a PUE of 1.10 across its fleet of data centers for the 12 months ending in Q1 2021,\footnote{\url{https://www.google.com/about/datacenters/efficiency/}}  compared to an average global PUE of 1.59~\cite{ascierto2020uptime}. 

Other factors, such as the emissions produced by maintenance workers driving to and from the data center, emissions from manufacturing the computer systems, and emissions from building the structure in which the data center is housed\footnote{One of the largest single source of \carbondioxide emissions, contributing to 7\%-8\% of global emissions, is the production of cement~\cite{iea-cement}.} are non-negligible but beyond the scope of this paper.
Finally, for workloads that do not use GPUs (e.g., storage or web hosting) we recommend users choose low emissions regions and times of day, as they will not have access to single-instance emissions calculations. 
We leave it open for future research to address how to appropriately allocate \carbondioxide emissions from such data center-wide processes to individual reserved cloud instances.

\section{Electricity consumption for AI Workloads}\label{sec:electricity}

\begin{table*}[h]
\centering
\begin{tabular}{ |c||c|c|c||c|c|c||c|c|c|c|c| } 
 \hline
 Model & BERT  & BERT  & 6B & Dense  & Dense  & Dense  & ViT  & ViT  & ViT  & ViT  & ViT \\
 & finetune & pretrain & Transf. &  121 &  169 &  201 &  Tiny &  Small &  Base &  Large &  Huge\\
 \hline
 GPU & 4$\cdot$V100 & 8$\cdot$V100 & 256$\cdot$A100 & 1$\cdot$P40 & 1$\cdot$P40 & 1$\cdot$P40 & 1$\cdot$ V100 & 1$\cdot$V100 & 1$\cdot$V100 & 4$\cdot$V100 & 4$\cdot$V100 \\
 \hline
 Hours & 6 & 36 & 192 & 0.3 & 0.3 & 0.4 & 19 & 19 & 21 & 90 & 216\\ 
 \hline
 kWh & 3.1 & 37.3 & 13,812.4 & 0.02 & 0.03 & 0.04 & 1.7 & 2.2 & 4.7 & 93.3 & 237.6\\ 
 \hline
\end{tabular}
\caption{For the 11 models in our analysis: the type of GPU, the number of GPUs of that type, the number of hours, and the energy used in kWh. For example, our BERT language modeling (BERT LM) experiment used 8 V100 GPUs for 36 hours and used a total of 37.3 kWh. We note our training run of the 6 billion parameter transformer only trained for approximately 13\% of the time it would take to train to completion, we estimate a full training run would consume approximately 103,593 kWh.}
\label{tab:gpu_hours}
\end{table*}

As outlined in \S\ref{sec:methodology}, calculating software carbon intensity begins with recording the electricity consumption, which can then be mapped to emissions based on the emissions of the grid being used. In this section, we present data on electricity consumption for experiments training 11 different models, covering natural language processing (NLP) and computer vision applications, ranging from less than an hour on a single GPU up to more than 8 days on 256 GPUs. We outline both the experiments themselves and their electricity consumption, and in the following section we use the electricity consumption and carbon intensity tool described in the previous section to calculate their software carbon intensity.

\subsection{NLP}

\paragraph{BERT Training} We monitored the energy consumption while training a BERT-small model~\cite{devlin2019bert} for approximately 36 hours on 8 NVIDIA V100 GPUs. That training run consumed over 37 kWh of electricity.

\paragraph{BERT Finetuning} We tracked the energy consumption while finetuning the BERT-small model on a standard natural language inference task~\citep[MNLI]{williams2017broad} for approximately 6 hours on 4 NVIDIA V100 GPUs. Our finetuning run consumed around 3.2 kWh of electricity, i.e.,  less than one tenth that due to BERT-small pre-training.

\paragraph{6 Billion Parameter Transformer} We tracked the energy consumption of training a large language model comprising over 6.1 billion parameters during 8 days on 256 NVIDIA A100s. The total energy amounted to a staggering 13.8 MWh. This model was not trained to completion, but only until 13\%; a full training run would take 60 days. Thus, we estimate the total energy consumption to train this model to completion would be approximately $(60/8)*13.8=103.5$ MWh, or 103,500 kWh --- almost 2800 times more than training the BERT-small model!

\subsection{Computer Vision}

\paragraph{DenseNets} We trained three sizes of DenseNets~\cite{densenet} on MNIST~\cite{lecun1998gradient}. The jobs lasted between 20 and 25 minutes and consumed between 20 and 38Wh (or 0.02 to 0.04 kWh) of electricity, which is negligible compared to the other models.

\paragraph{Vision Transformers} We evaluated the energy consumption during the training of five sizes of Vision Transformers~\cite{vit} on ImageNet~\cite{deng2009imagenet}. For the smallest ViT experiment (ViT tiny), training lasted around 19 hours on a single V100 and consumed approximately 1.7 kWh. For the largest one (ViT huge), training lasted more than 9 days on a 4 V100s and consumed approximately 237 kWh. The full list of models can be found in Table \ref{tab:gpu_hours}.

\section{Emissions by Region and Time of day}
Using the methodology presented above, we provide some of the first measurements of the differences of actual datacenters from a major cloud provider. Importantly, what we have is a time series of marginal emissions: for example, if a job were to run from 1 pm to 5 pm in the US West region with a cloud instance that has four fully-utilized GPUs, both the energy consumed and the marginal carbon intensity during that time is what we want to record. This time-series data can estimate the cumulative emissions for that experiment at the end.

\begin{figure*}[h!]
    \centering
    \includegraphics[width=150mm]{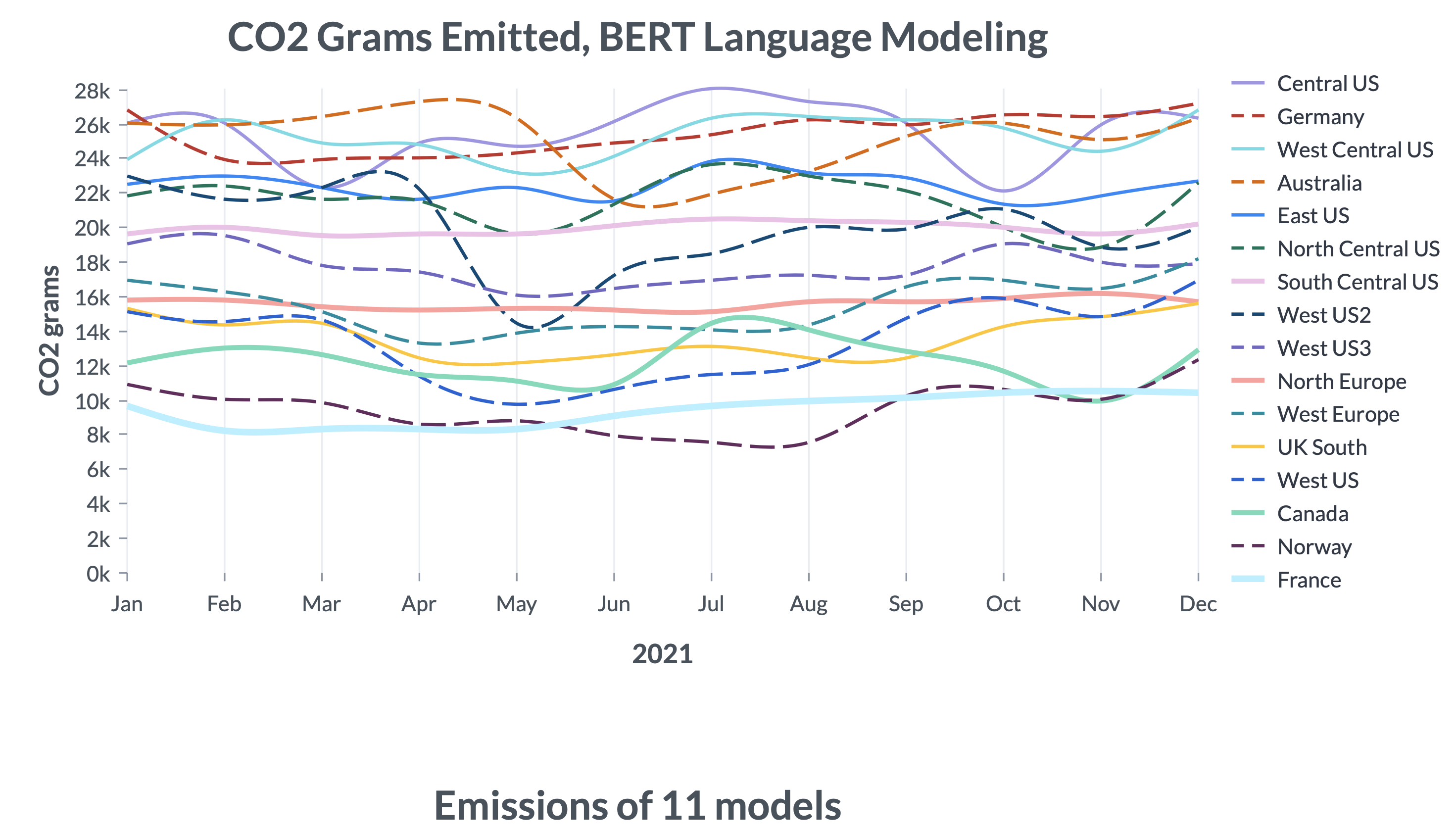}
    \caption{Carbon emissions that would be emitted from training BERT (language modeling on 8 V100s for ~36 hours) in 16 different regions (one region per line) at different times throughout the year. Each line is relatively flat, indicating the emissions in a single region during different months are relatively similar. There is large variation between the least carbon-intensive regions (the lowest lines) compared to the most carbon-intensive regions (the top lines), indicating that choosing the region in which experiments run can be very impactful (~7k grams vs.~26k grams, for the most efficient vs.~least efficient regions).}
    \label{fig:BERT_elec_load_by_region}
\end{figure*}

\subsection{Region}

How much does the choice of datacenter region impact the emissions? And for a single region, how much variation occurs throughout the year? We address these questions in Figure~\ref{fig:BERT_elec_load_by_region}, which shows carbon emissions that would be emitted from training BERT (see \S\ref{sec:electricity} for more details) on 8 V100 GPUs for 36 hours in 16 different regions (one region per line) at different times throughout the year.

\begin{figure*}[h!]
    \centering
    \includegraphics[width=150mm]{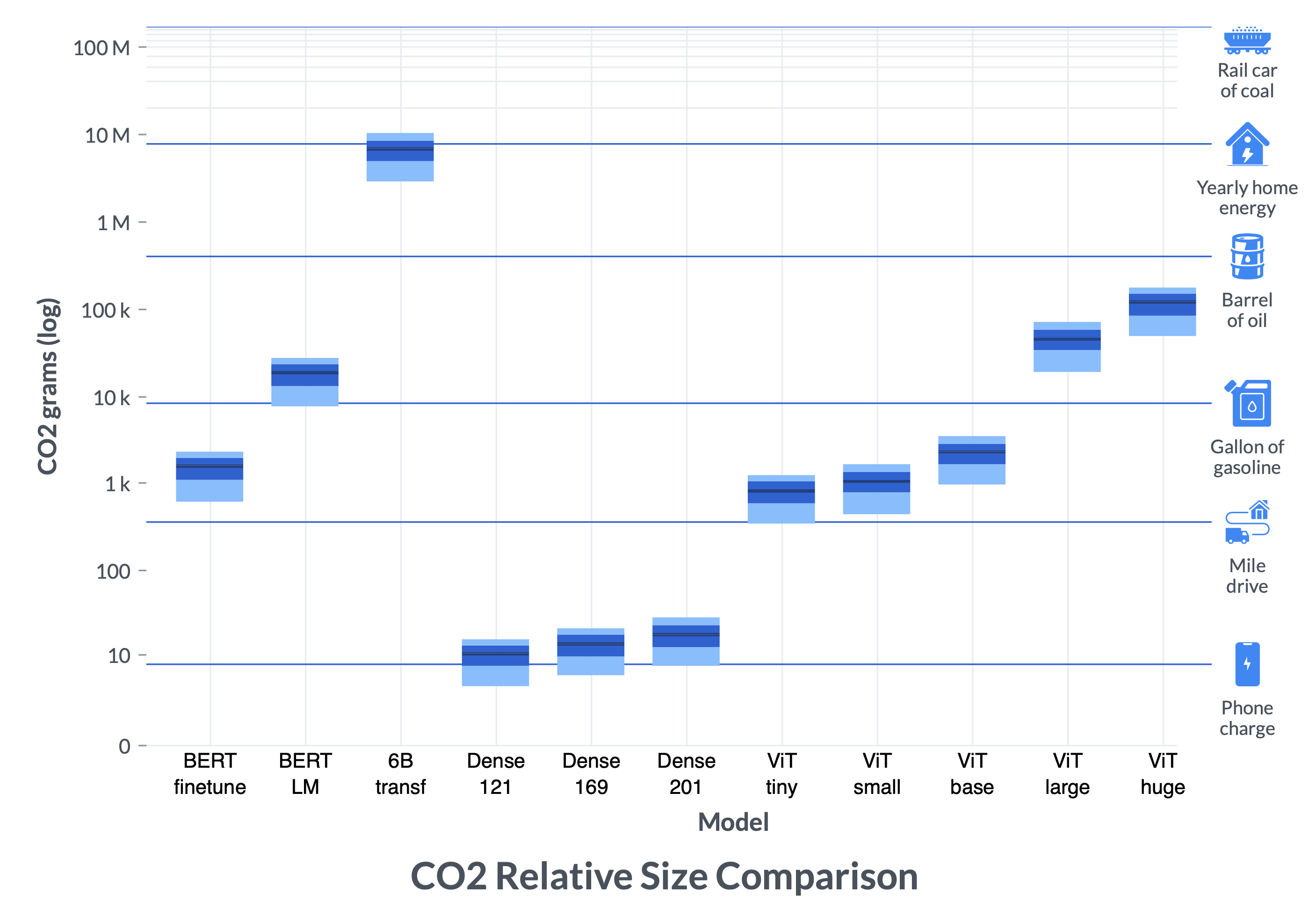}
    \caption{Emissions for our 11 experiments described in \S\ref{sec:electricity}. For each model we show a vertical blue bar, where the top of the bar is the max, the bottom is the min, and the black line represents the average emissions (across regions and time of year). First and fourth quartiles are represented by the light blue at the top and bottom of each vertical blue bar. The largest training runs (e.g., 6 billion parameter LM) releases a significant amount of emissions, no matter the region (and recall the 6 billion parameter LM is only trained for ~13\% of a full run, so a full run would emit about an order of magnitude more emissions than reported here). The smallest experiments emit very little. Presented on a log scale, with references on the right indicating equivalent sources of emissions per the \citet{epa_equivalences}.}
    \label{fig:all_elec_load_by_region}
\end{figure*}

What do emissions look like across the 11 experiments described in \S\ref{sec:electricity}? In Figure~\ref{fig:all_elec_load_by_region} we show results for all 11 experiments, which cover two BERT experiments (finetuning and language modeling), partial training of a 6.1 billion parameter Transformer, 3 sizes of DenseNets, and five sizes of Vision Transformers. Each experiment is represented by a vertical blue bar showing the range of emissions that would be emitted for that experiment across different regions. The top of the blue bar is the emissions from running that experiment in the region with the most emissions, the bottom is the emissions from running that experiment in the region with the least emissions, the black line represents the average, and the light blue regions are the top and bottom quartiles.

In Figure~\ref{fig:all_elec_load_by_region} we also include estimates of equivalent sources of emissions per the \citet{epa_equivalences}.
One phone charge is estimated to emit $8.22 \times 10^{-6}$ metric tons (using US national weighted average \carbondioxide marginal emission rate for delivered electricity), one mile driven is estimated to emit $3.98 \times 10^{-4}$ metric tons (using average US passenger vehicle, which gets 22.5 miles per gallon of gasoline), one gallon of gasoline consumed is estimated to emit $8.887 \times 10^{-3}$ metric tons, one barrel of crude oil consumed is estimated to emit $0.43$ metric tons, one average US home energy use is estimated to emit $8.30$ metric tons (using the sum of emissions from generating electricity, natural gas, liquid petroleum, and fuel oil), and one rail car of coal is estimated to emit $181.29$ metric tons.

The largest experiment in our set is the 6 billion parameter transformer, and that model is only partially trained (as described in \S\ref{sec:electricity}, it is only trained for about 13\% of the time needed to converge). Even partially trained, experiments of this size can emit more \carbondioxide than all emissions from the average US home for a year (which includes emissions from electricity generation, natural gas, liquid petroleum gas, and fuel oil, totaling 8.3 metric tons \carbondioxide per year). Perhaps unsurprisingly, even the most efficient region of those we examined for that experiment still leads to more emissions than a full barrel of oil. If this had been trained to completion, we estimate it would have emitted 21 to 78 metric tons of \carbondioxide (depending on the region it was run in).


Comparing against previous work on measuring emissions can be challenging without full information about data and model parallelism, GPU utilization, the number of weight updates, and other relevant factors; while we don't have experiments covering the same models as previous work on estimating \carbondioxide, we can make approximate comparisons along three dimensions: a) kWh per GPU hour, b) \carbondioxide grams per GPU hour, and c) \carbondioxide grams per kWh.
Here we compare against \cite{patterson2021carbon} and \cite{patterson2022carbon} which report information about training especially large models.
Their estimates also include additional sources of \carbondioxide, like PUE (Power Usage Effectiveness) of their datacenters, so we expect their kWh per GPU hour and \carbondioxide per GPU hour to be higher than our estimates (which only count the GPU electricity consumption).

Across our experiments, we find kWh per GPU hour to range from 0.07 to 0.28, compared to \citet{patterson2021carbon} with 0.22 to 0.47, and \citet{patterson2022carbon} with 0.36.
We find \carbondioxide (grams) per GPU hour in the most efficient region to average 34, and in the least efficient region to average 128, where \citet{patterson2021carbon} found a range of 63 to 202, and \citet{patterson2022carbon} found 32.
We find \carbondioxide (grams) per kWh in the most efficient region to average 200, and in the least efficient region to average 755. The estimates from \citet{patterson2021carbon} range between 427 and 545 (except GShard 600B with 200), and \citet{patterson2022carbon} found 88. In short, we find most of their estimates to be within the range of ours, with the exception of \citet{patterson2022carbon}, which specifically aimed to choose a region that was more \carbondioxide efficient.

\begin{table*}[htbp!]
\centering
\begin{tabular}{ |c||c||c||c|c|c|c|c|c|c|c|} 
 \hline
  & & Hour & 0:00 & 03:00 & 06:00 & 09:00 & 12:00 & 15:00 & 18:00 & 21:00 \\
 \cline{3-11}
BERT & Central & Day 1 & 2,381 & 2,341 & 2,210 & 2,252 & 2,354 & 2,391 & 2,410 & 2,403 \\
 \cline{3-11}
 finetune & US & Day 2 & 2,330
 & 2,249 & 2,204 & 2,299 & 2,320 & 2,317 & 2,339 & 2,344 \\ 
 \cline{3-11}
 &  & Day 3& 2,430 & 2,339 & 2,257 & 2,313 & 2,393 & 2,374 & 2,317 & 2,331 \\ 
 \hline
\end{tabular}
\caption{How do emissions vary throughout different times of day? We present the emissions produced by the BERT finetuning experiment described in \S\ref{sec:electricity} had it run at different times in the Central US region, on three separate days.}
\label{tab:time_of_day}
\end{table*}

\subsection{Time of Day}

While the choice of region is a major source of variation in \carbondioxide emissions, diurnal variations also play a significant role. During the day, a region may have a higher mix of renewable energy or fossil-fuel based source~\cite{de2019100}. As one can see in Table~\ref{tab:time_of_day}, depending on the day, starting the BERT finetuning at, e.g., midnight instead of 6:00 can result in carbon emissions increasing by up to 8\%. The amount of variation varies by region and time of year as\break well.

\section{Optimizing Cloud Workloads}

\begin{figure*}[ht!]
    \centering
    \subfloat[\textit{Flexible Start} optimization for Dense 201.]{\includegraphics[width=75mm]{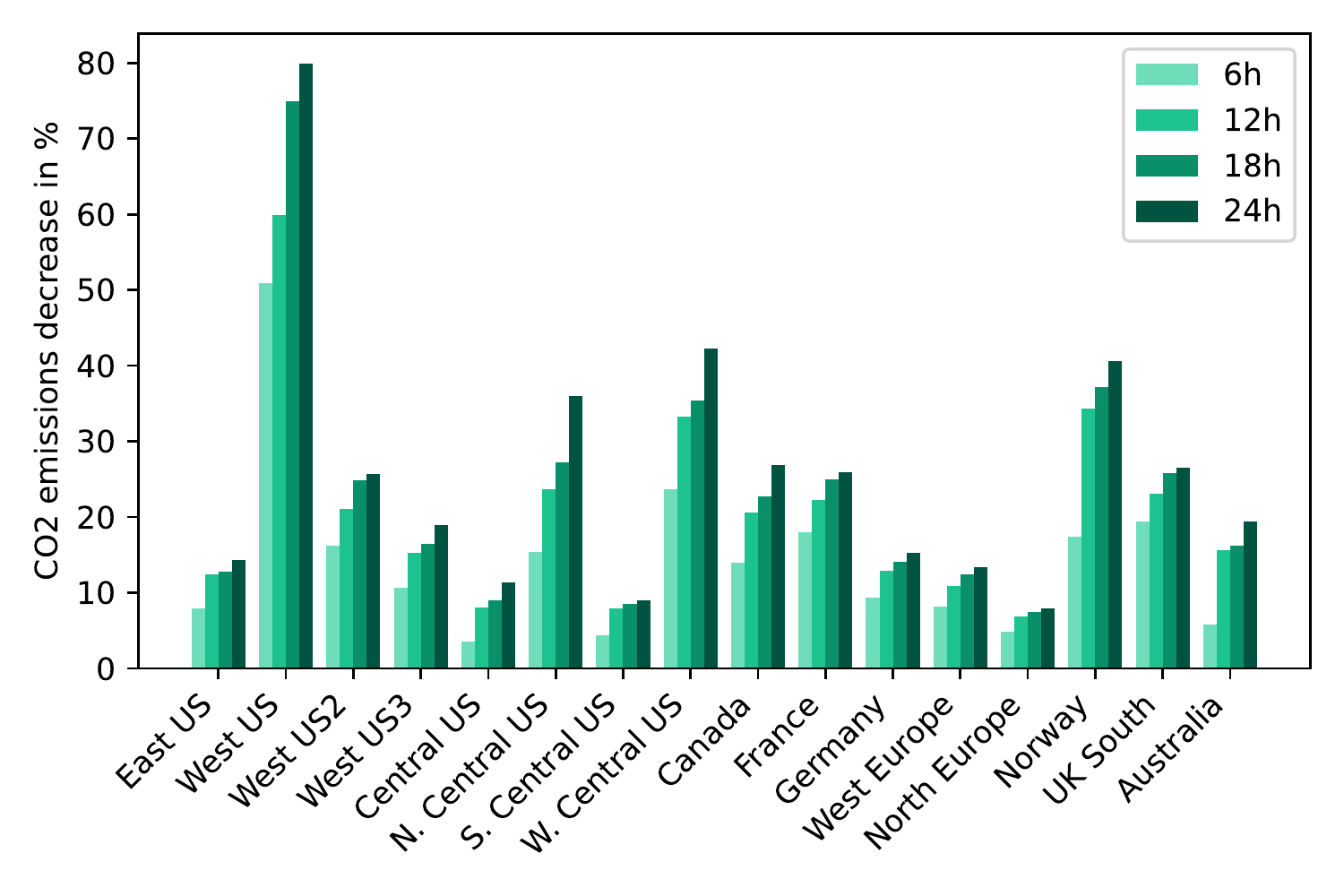}}
    \subfloat[\textit{Flexible Start} optimization for 6B parameters Transformer.]{\includegraphics[width=75mm]{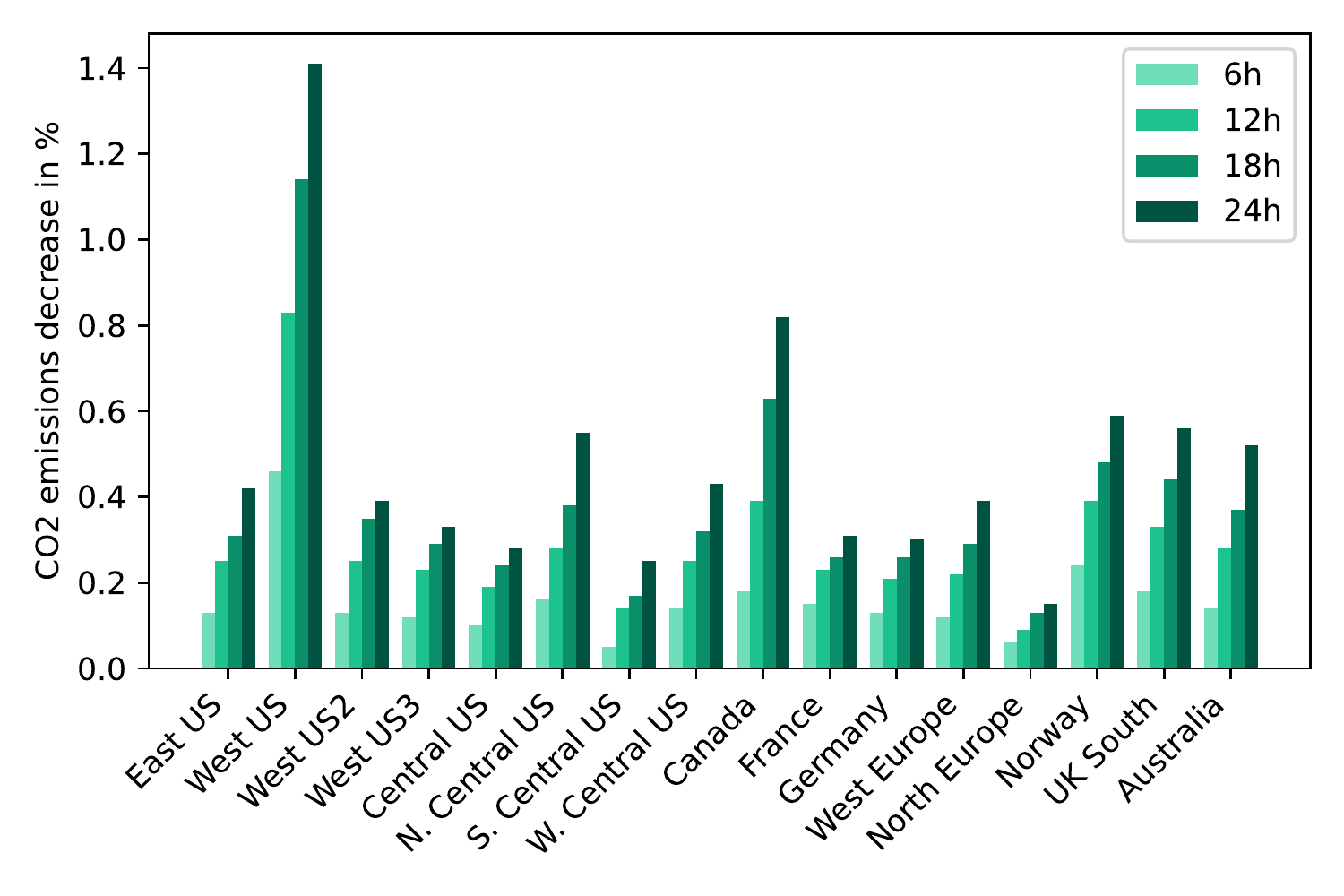}}
    \caption{What proportion of emissions can we expect to save if we change the start time by up to 24 hours? For very short experiments like DenseNet 201 (a), which ran for less than half an hour, we can find significant reduction, greater than 30\% in multiple regions, and up to 80\% in West US; for very long runs like training a 6 billion parameter language model for 8 days (b), changing the start time by up to 24 hours leads to less than 1.5\% reduction at best in any region. Note: we confirmed with WattTime that emissions estimates for West US were correct, that region has large variance.}
    \label{fig:start_optimizations_main_text}
\end{figure*}

We use the tools presented so far to evaluate two algorithms for reducing emissions of AI workloads on the Microsoft Azure cloud compute platform using temporal shifting. We consider sixteen regions where workloads can be scheduled on Azure: nine in North America, six in Europe and one in Australia (see Figure~\ref{fig:start_optimizations_main_text}). For each region, we obtained from WattTime the historical marginal carbon emissions for the year 2020 at a 5-minute granularity. We also measured the electricity consumption per 5-minute intervals of the various models introduced in \S\ref{sec:electricity}. The two optimization methods we studied are:


\begin{itemize}
    \item \textit{Flexible Start}. Start the workload at the time, in the next $N$ hours, that minimizes its carbon emissions. Once the workload is launched, it is run until completion. Implementation: Consider all possible start times (in 5 minute increments) in the desired window. For each start time, compute the job's corresponding emissions and pick the lowest.
    
    \item \textit{Pause and Resume}. Assuming the workload can be stopped and restarted (a fairly weak constraint), run its computations over the next $(N + \mbox{job duration})$ hours while minimizing its total carbon emissions. This involves pausing and resuming the job, possibly multiple times, to avoid consuming energy when carbon intensity is high. Implementation: Find the 5 minute intervals with the lowest marginal emissions during the $(N + \mbox{job duration})$ hour window, and select enough intervals to add up to the job duration. Then simulate running the job only during those intervals and compute the corresponding emissions. We explored two sets of values for $N$: one absolute, corresponding to increasing the total duration of the job by at most \{6, 12, 18, 24\} hours; and a second one relative, where we allow the job to increase in duration by at most \{25\%, 50\%, 75\%, 100\%\}. In other words, for the second set, we allow the workload to last for at most twice its duration had it not been stopped. While arbitrary, we motivate the choice of those two sets by the extreme range of possible job duration (from minutes to weeks). Note that we assume pausing and restarting the job is immediate and does not consume additional energy: this is similar in spirit (for carbon emissions) to Spot Instances on existing cloud platforms which automatically pause an instance if its price rises above a threshold set by the user.
\end{itemize}

We find the region that the algorithms are evaluated in has a significant impact. For example, the region we labeled West US varies frequently throughout a single day between periods of high emissions and very low emissions, and thus \textit{Pause and Resume} can lead to significant reductions. However, other regions do not present as much variance, and thus lead to less reduction in emissions. See Figures~\ref{fig:start_optimizations_main_text} and \ref{fig:pause_resume_optimizations_main_text}. The lack of geographic diversity in the region list is an unfortunate consequence of the unavailability of carbon intensity data from other continents; we hope such data becomes broadly available in the near future.

\subsection{Evaluation of Emissions Reduction Algorithms}

We evaluate how the two optimization algorithms would impact the emissions from the 11 experiments described in \S\ref{sec:electricity}. In order to account for daily variations (weather, electricity demand, etc.), we report the average emissions decrease computed over 5 different start times in each month, giving a total of 60 data points.

\subsubsection{Emissions Reduction by Region}

\paragraph{Flexible Start}
When evaluating the \textit{Flexible Start} algorithm for a fixed duration between 6 hours and 24 hours, we find significant emissions reductions for shorter jobs (e.g., the DenseNet experiments), with minimal savings for jobs that are longer than a day; this aligns with our expectations, as short jobs can be run when emissions are lowest throughout a day, but long jobs naturally average across multiple days. See Figure~\ref{fig:start_optimizations_main_text}, with results for all experiments in the appendix. This analysis is designed to highlight a use case where an AI workload needs to run regularly, but the practitioner has some flexibility on when it runs (so it could, e.g., run over night, if that is when carbon intensity is lowest). This is in fact a common use case in production ML systems deployed at companies, where models are re-trained on a regular schedule to incorporate new data over time \cite{hazelwood2018hpca}.

\paragraph{Pause and Resume}
When evaluating the \textit{Pause and Resume} algorithm for durations up to 100\% of the duration of the original experiment, we find the opposite of the \textit{Flexible Start} result: short experiments like DenseNet 201 only see emissions reductions smaller than 10\%, while the 6 billion transformer training run (our experiment with the largest carbon intensity) actually sees the largest decrease in emissions. See Figure~\ref{fig:pause_resume_optimizations_main_text} for two examples, with results for all 11 experiments in the appendix. This analysis is designed to highlight a use case where an AI workload can be increased in duration by some proportion of the original run time.

\begin{figure*}[ht!]
    \centering
    \subfloat[\textit{Pause and Resume} optimization for Dense 201.]{\includegraphics[width=75mm]{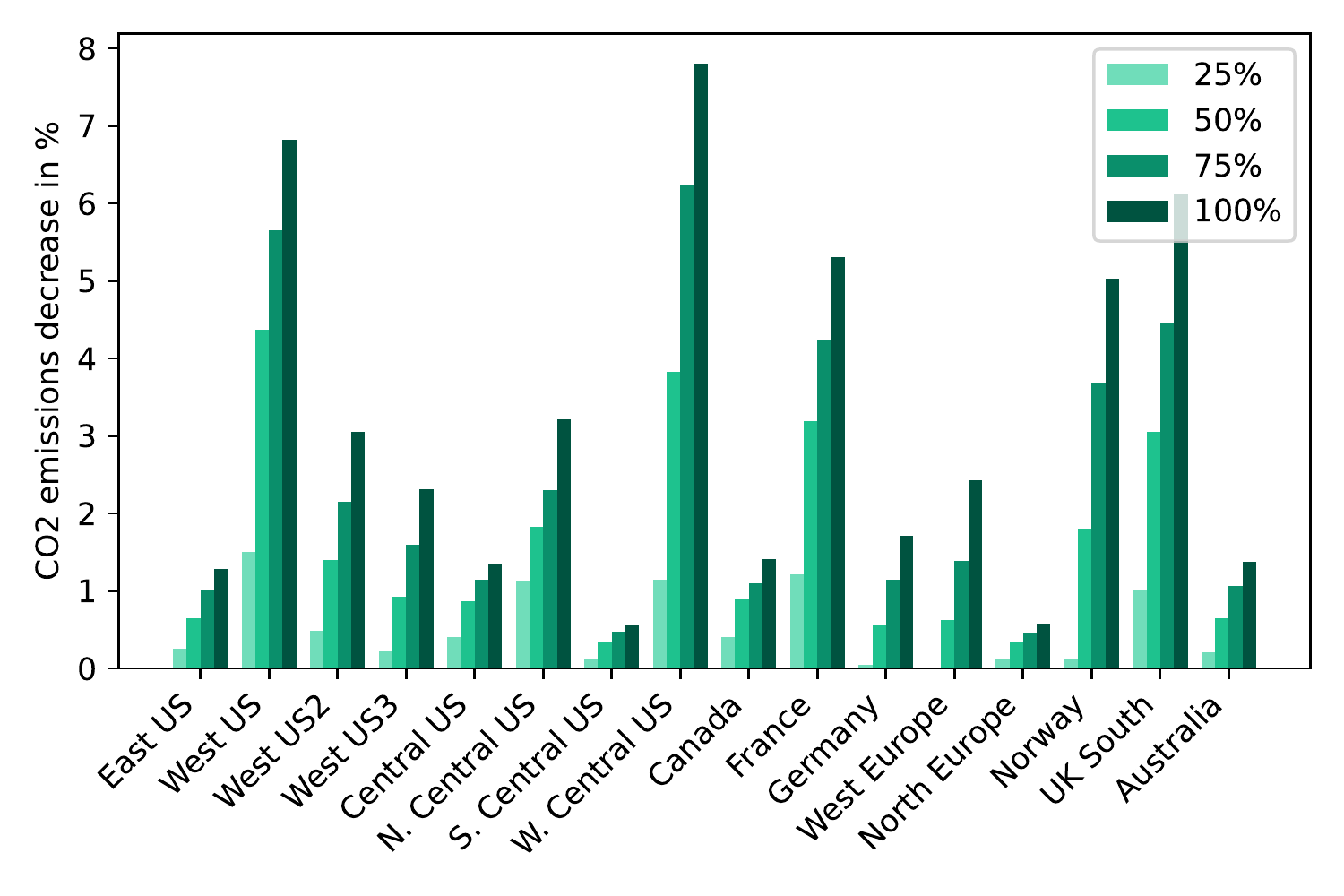}}
    \subfloat[\textit{Pause and Resume} optimization for 6B parameters Transformer.]{\includegraphics[width=75mm]{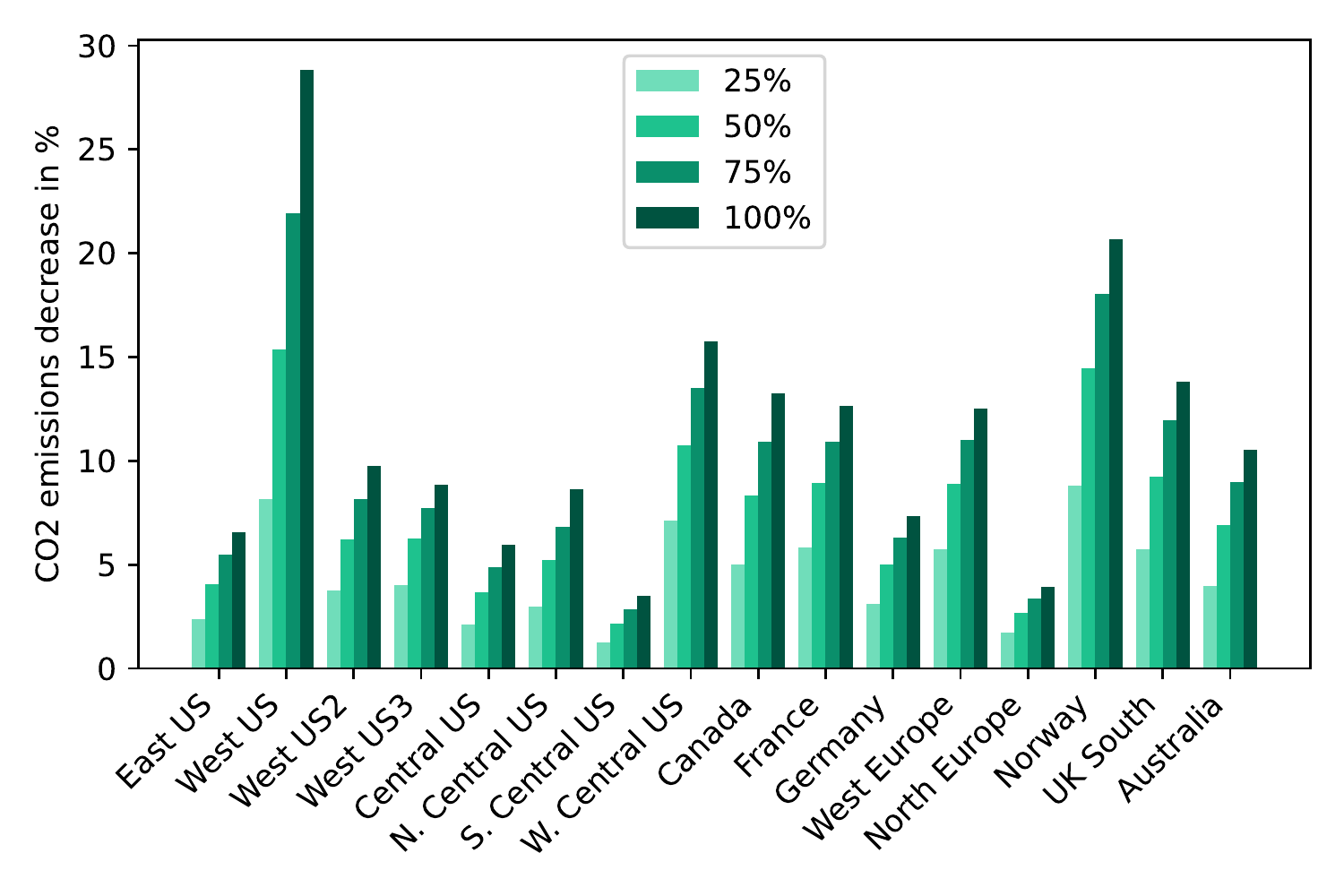}}
    \caption{What proportion of emissions can we expect to save if we pause an AI workload when emissions in a region are high and resume when emissions are low, increasing the total duration by up to double the original duration? For short experiments, the doubled duration is still relatively short, and thus leads to minimal emissions reduction (see DenseNet 201 in (a)); for very long runs like our 6 billion parameter language model training run in (b), which ran for 8 days, doubling the duration can lead to significant savings up to about 25\%. We confirmed with WattTime that emissions estimates for West US were correct, as that region has large variance.}
    \label{fig:pause_resume_optimizations_main_text}
\end{figure*}

\subsubsection{Comparable Duration Increases}
In the previous section we examined the amount of emissions reduction for our two algorithms by region, and compared \textit{Pause and Resume} increasing duration by a proportion of the original experiment and \textit{Flexible Start} by a fixed duration.
Here we evaluate the two algorithms when they increase the duration of an AI workload by the same amount (each result is averaged across all regions and times of year). One can think of the \textit{Flexible Start} algorithm as a version of \textit{Pause and Resume} where there is only one start time, and no pausing allowed; thus we should expect the \textit{Flexible Start} results to always lower bound the \textit{Pause and Resume} ones. 

We show results for both algorithms and two situations: increasing the duration of the run by 24 hours in Table~\ref{tab:6_main}, and by 100\% in Table~\ref{tab:100_main}. In these tables we also include information about the average number of pauses per hour for the \textit{Pause and Resume} algorithm. Perhaps surprisingly, we find the average number of pauses is quite low. This can be interpreted as the number of times the carbon intensity crosses above the threshold minimizing total emissions being small.

\begin{table*}[ht!]
\centering
\begin{tabular}{ | c||c|c|c||c|c|c||c|c|c|c|c | } 
\hline
Model & BERT  & BERT  & 6B & Dense  & Dense  & Dense  & ViT  & ViT  & ViT  & ViT  & ViT \\
 & finetune & LM & Transf. & 121 & 169 & 201 & Tiny & Small & Base & Large & Huge \\
\hline
\hline
FS & 14.5\% & 3.4\% & 0.5\% & 26.8\% & 26.4\% & 25.9\% & 5.6\% & 5.3\% & 4.2\% & 1.3\% & 0.5\% \\
\hline
\hline
P\&R  & 19.0\% & 8.5\% & 2.5\% & 27.7\% & 27.3\% & 27.1\% & 12.5\% & 12.3\% & 11.7\% & 4.7\% & 2.4\% \\
\hline
Pauses  / hr & 0.23 & 0.3 & 0.15 & 0.06 & 0.07 & 0.08 & 0.3 & 0.3 & 0.3 & 0.23 & 0.14 \\
\hline
\end{tabular}
\caption{For the 11 models in our analysis: the gain in percent averaged over the year and across the 16 regions for the \textit{Flexible Start} (FS) and \textit{Pause and Resume} (P\&R) optimizations allowing for a \textbf{24h increase} in job duration. The last line represents the average number of pauses per hour performed by the P\&R optimization.}
\label{tab:6_main}
\end{table*}

Note that the above optimizations were performed using historical data, meaning that their results are the best achievable, assuming access to an oracle predicting carbon intensity perfectly. \href{https://www.watttime.org/api-documentation/#emissions-forecast}{WattTime} currently provides marginal emission rate estimates and forecasts for up to 24 hours, so for short workloads, our findings will reflect gains observed in practice using the forecasts. For longer workloads, our numbers give an upper bound on the realizable gains. 
For example, the \textit{Pause and Resume} algorithm pauses the workload when emissions are above a threshold, and resumes when emissions are below that threshold. In our evaluation here we set this threshold such that the total run time is increased by, e.g., 24 hours; a machine learning practitioner would have to estimate how much a particular threshold would increase job duration, but would not know exactly.
The dynamic nature of the \textit{Pause and Resume} optimizations suggests that well-designed scheduling algorithms should be able to get rather close to the upper-bound. We leave such algorithms to future work and hope our tools can inspire further research into that type of scheduling. Moreover, it is likely that carbon intensity forecasting will improve over the years and eventually extend beyond 24 hours, allowing time-shifting decisions to become increasingly accurate.

\begin{table*}[ht!]
\centering
\begin{tabular}{ | c||c|c|c||c|c|c||c|c|c|c|c | } 
\hline
Model & BERT  & BERT  & 6B & Dense  & Dense  & Dense  & ViT  & ViT  & ViT  & ViT  & ViT \\
 & finetune & LM & Transf. & 121 & 169 & 201 & Tiny & Small & Base & Large & Huge \\
\hline
\hline
FS & 7.0\% & 4.1\% & 2.6\% & 1.8\% & 2.5\% & 2.7\% & 5.0\% & 4.8\% & 3.9\% & 3.3\% & 3.0\% \\
\hline
\hline
P\&R  & 9.5\% & 11.0\% & 11.4\% & 2.0\% & 2.8\% & 3.1\% & 11.0\% & 11.0\% & 10.8\% & 11.4\% & 11.3\% \\
\hline
Pauses  / hr & 0.42 & 0.29 & 0.27 & 1.5 & 1.88 & 2.0 & 0.31 & 0.32 & 0.31 & 0.27 & 0.26 \\
\hline
\end{tabular}
\caption{For the 11 models in our analysis: the gain in percent averaged over the year and across the 16 regions for the \textit{Flexible Start} (FS) and \textit{Pause and Resume} (P\&R) optimizations allowing for a \textbf{100\% increase} in job duration. The last line represents the average number of pauses per hour performed by the P\&R optimization.}
\label{tab:100_main}
\end{table*}

\section{Considerations for Model Development and Deployment \label{sec:suggested}}

Generally speaking, we advocate that researchers and practitioners record and report the amount of emissions incurred by ML projects, starting with the initial exploratory training phases all the way through hyperparameter tuning and deployment for the final model. This can inform an Operational Lifecycle Analysis (OLCA) for a machine learning model, which would account for all phases of the ML lifecycle. In the subsections below, we outline some ways in which the proposed tool can be used at different stages of the model development and deployment process, and describe some environmental impacts due to ML modeling that are outside the scope of measurement of this tool.



%


We see various ways in which our tool can help guide model training, for instance via carbon-informed optimization (similarly to what ~\cite{kim2021autofl} proposed for energy efficiency in federated learning), or for cloud-based recommendations that enable users to opt-in for carbon-aware configurations (in terms of region, time, etc.) to reduce the carbon intensity of their training workloads. We believe that tracking and reducing greenhouse gas emissions can be a very important feature for users deciding on how they will set up their cloud usage, but we recognize that there are natural trade-offs that must be considered. We therefore recommend that the measurements provided by our tool be used to guide informed decisions alongside other considerations as part of a holistic approach, and not as a single gold standard to optimize against.  For example, even just within the scope of ML model development, it often takes engineering time to optimize a workload to be more efficient (i.e., use less computational resources), and a user should consider whether that time would be better spent elsewhere (e.g., transferring the workload to another region with lower average emissions).  Furthermore, some projects have strict time constraints, and so scheduling jobs to only run at night would significantly delay progress, potentially leading to more emissions in other parts of the project. Thus, our suggestions are not meant as a one-size-fits-all solution which will eliminate carbon emissions, but instead as a set of options which can be referenced by users and decided upon on a case-by-case basis.  Finally, there are also many additional upstream and downstream emissions considerations due to the ML model lifecycle, due to, e.g., hardware manufacturing and downstream uses or misuses of the model, that could eclipse the direct emissions due to model training alone. See \S\ref{sec:related-work} for further discussion of this crucial point.

Another important consideration is operating cost; it could be the case that Region A is lower emissions but higher cost than Region B for a particular workload, and thus a user could run their workload in Region B and have some budget left over that could be used for other reductions in emissions. A final consideration is cost of data transfer; it could be the case that Region A is lower emissions and monetary cost than Region B for a particular workload, but the energetic, environmental, or monetary cost of moving the data could exceed the benefits gained.

If we see broad adoption of such reporting tools, we may see increases in cloud use in regions which have low emissions.
In such a scenario, providers could be incentivized to build new data centers, and providers should consider the local impact of such construction.

\section{Future Directions \label{sec:future}}

As mentioned in \S\ref{sec:suggested}, single-instance emissions are a well-defined starting place for quantifying, mitigating, and reducing the environmental impact due to ML, but do not present a complete picture of the total emissions that should be accounted for when considering the overall carbon emissions of the ML life cycle. Here are some aspects that are yet to be accounted for (and in some cases, yet to be defined) in terms of the overall OLCA of machine learning:

\paragraph{Scopes of emissions} The \href{https://ghgprotocol.org/}{Greenhouse Gas Protocol (GHGP)} is a standard created by the World Resources Institute and the Business Council for Sustainable Development, and has seen broad adoption internationally. It defines Scope 1, Scope 2, and Scope 3 emissions as follows: Scope 1 emissions are those generated by direct actions of a company, such as running motor vehicles; Scope 2 emissions are those associated with purchase of electricity, steam, heating, or cooling; and Scope 3 emissions are those that the company indirectly participates in, such as those due to investments of the company and downstream use of products. In the present work, we have focused on the Scope 2 emissions incurred due to electricity usage by cloud providers. The current GHGP Scope 2 is an attributional guidance that precludes the use of marginal emissions rates, and primarily focuses on broad generation-based average rates. It is important to note that the GHGP Scope 2 guidance is incompatible with the proposed method; this paper illustrates the need to revisit the Scope 2 guidance to better align with consequential accounting methods.  

We do not cover the Scope 1 emissions (e.g. emissions that directly result from business activities, such as stationary combustion of fuels for backup power generation in cloud datacenters), for a more detailed discussion see e.g. \citet{gupta2021chasing}, nor the Scope 3 emissions (e.g. emissions that indirectly result from all other business activities, such as those associated with the upstream raw materials extraction, manufacturing, and delivery of cloud-based IT asset infrastructure such as servers from suppliers to be used in a cloud provider's datacenters). Both of these types of emissions warrant discussion and debate by the AI community--- and indeed some work has begun on the subject, e.g.,~\cite{kaack2021aligning,ligozat2021unraveling}---but we are missing a more concrete structure for categorizing, quantifying and mitigating the different scopes of emissions in our field. This would involve the active participation of specific stakeholders to establish the tooling and reporting required to better estimate these aspects, which is a challenge in itself.

\paragraph{Developing certification systems for ``Green AI''} While initiatives like the Green Software Foundation are making important progress towards measuring and mitigating the carbon footprint of software in general, the decentralized and data-driven nature of ML will call for specific approaches and guidelines to ensure its efficiency. We anticipate that AI-specific initiatives, spanning both research and academia, will help establish certification systems (or badge systems) that will allow both model developers and users make more informed choices with regards to sustainability. The current framing of Scopes 1, 2, and 3 may not encompass all the emissions reasonably associated with an AI program.

\paragraph{Improving the carbon transparency of research and practice} Despite the existence of tools such as Code Carbon~\cite{schmidt2021codecarbon} and EvergyVis~\cite{shaikh21energyviz}, both carbon estimation and reporting in ML publications and technical reports remain a relatively rare phenomenon. Conferences such as NeurIPS and NAACL have recently added emissions reporting as an optional part of the submission process; however, more encouragement will be necessary for this to become commonplace. Gathering more data about the environmental impact of our field is a crucial step towards identifying room for improvement and, eventually, reducing our emissions. 

\paragraph{Supporting improved estimates of emissions rates} The estimates of emissions rates providers would benefit from more and better data being provided by electric system operators. This is particularly true in areas of the world where it is currently not possible to produce hourly marginal estimates.  

\paragraph{Reducing AI's scope-enabled emissions} Responsible development and application of AI must account not only for the hidden costs of development, as discussed in this paper, but for the positive or negative carbon impact the application enables. AI models continue to be used for oil exploration~\cite{nordloh2020machine}, deforestation~\cite{mosin2019remote}, and mining~\cite{hyder2019artificial}, among other environmentally-detrimental practices. When considering the net impacts of an AI application, it is imperative to determine the extent to which AI is incentivizing practices that have a negative impact on the environment, or the extent to which applications are directly reducing emissions or otherwise incentivizing practices that are beneficial to the climate, and take these downstream direct and indirect effects into account in the overall environmental impact assessment of our field~\cite{birhane2021values, kaack2021aligning}.


\begin{acks}
We thank Avi Allison (Microsoft) for insights associated with carbon accounting and Location-based Marginal Emissions (LME) data, 
Henry Richardson (WattTime) for insights on LME data and the Software Carbon Intensity (SCI) specification, Abhishek Gupta for his work on the SCI specification, and
Ananya Ganesh (CU Boulder) for help in obtaining the measurements included in Table~\ref{tab:gpu_vs_other}. We also thank Alessandro Sordoni, Payal Bajaj,  and Vibhav Vineet for sharing their training and inference jobs, and Jon Borchardt for help with plotting.
\end{acks}


\bibliographystyle{ACM-Reference-Format}
\bibliography{azure}

\onecolumn
\appendix

\section{Additional Plots}

\begin{figure}[ht]
    \centering
    \subfloat[\textit{Flexible Start} optimization.]{\includegraphics[width=75mm]{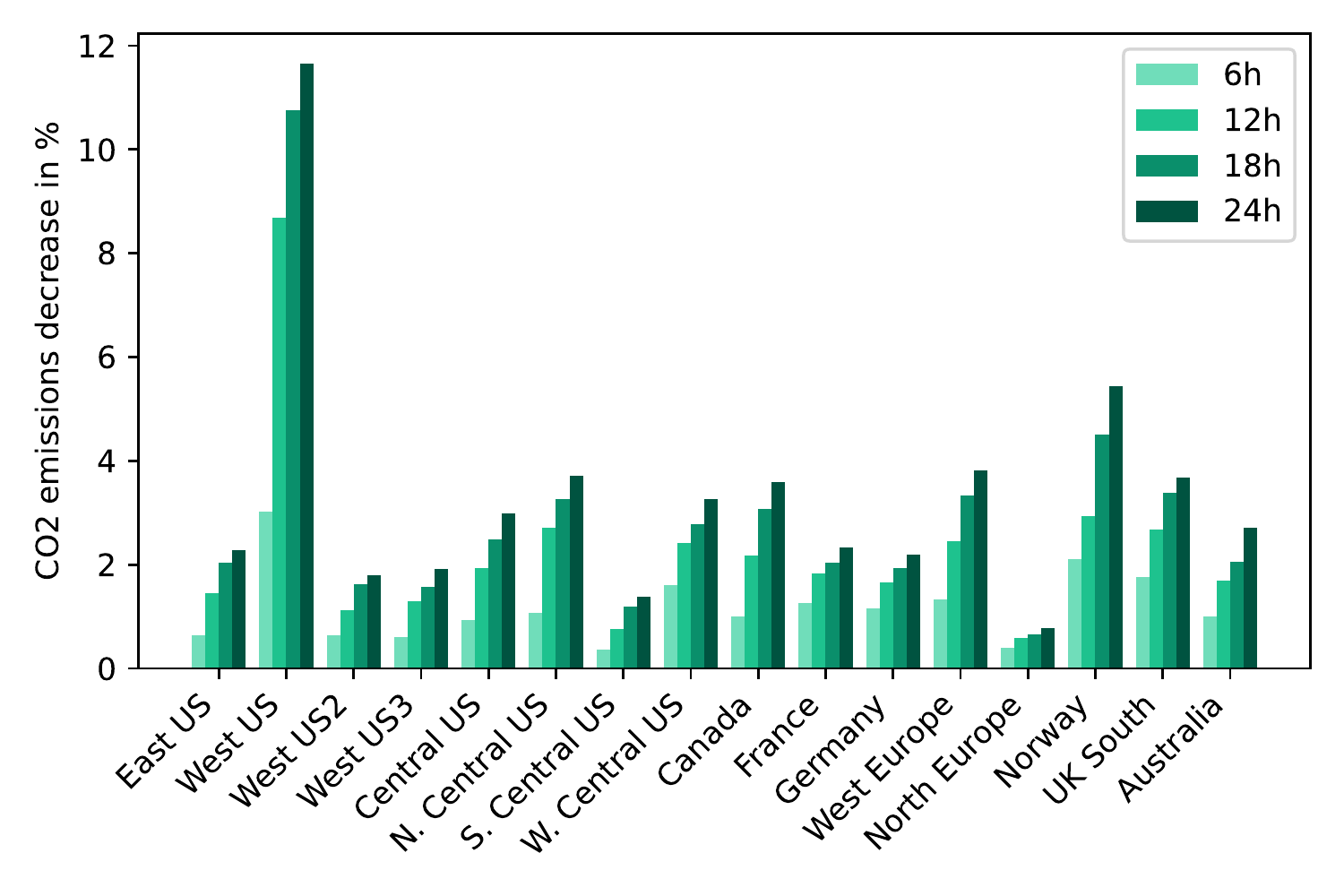}}
    \subfloat[\textit{Pause and Resume} optimization.]{\includegraphics[width=75mm]{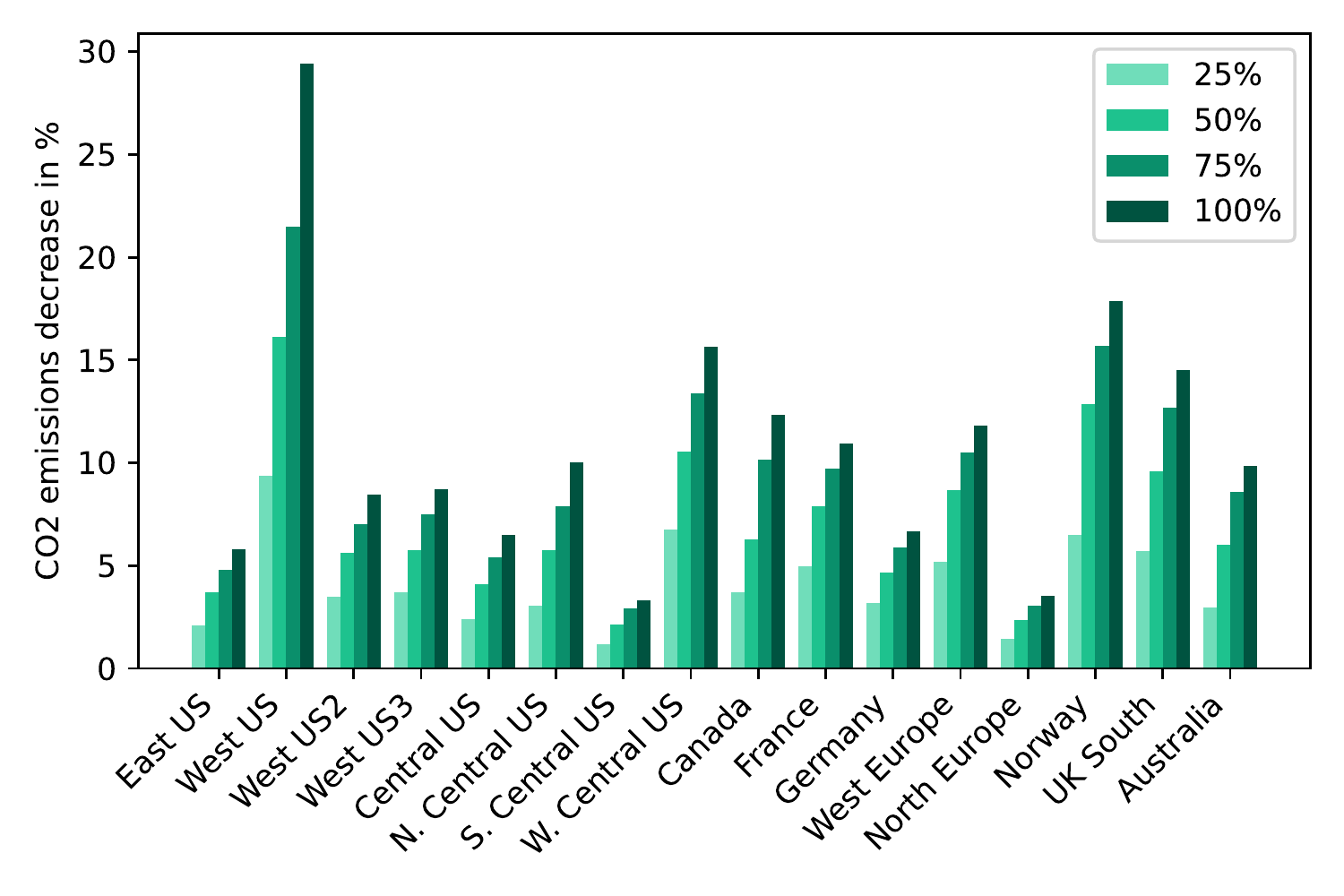}}
    \caption{Optimization results for the training of BERT small on 8 V100s. Without optimization, the job ran for approximately 36 hours and consumed 37.3 kWh.}
    \label{fig:bert_training}
\end{figure}

\begin{figure}[h]
    \centering
    \subfloat[\textit{Flexible Start} optimization.]{\includegraphics[width=75mm]{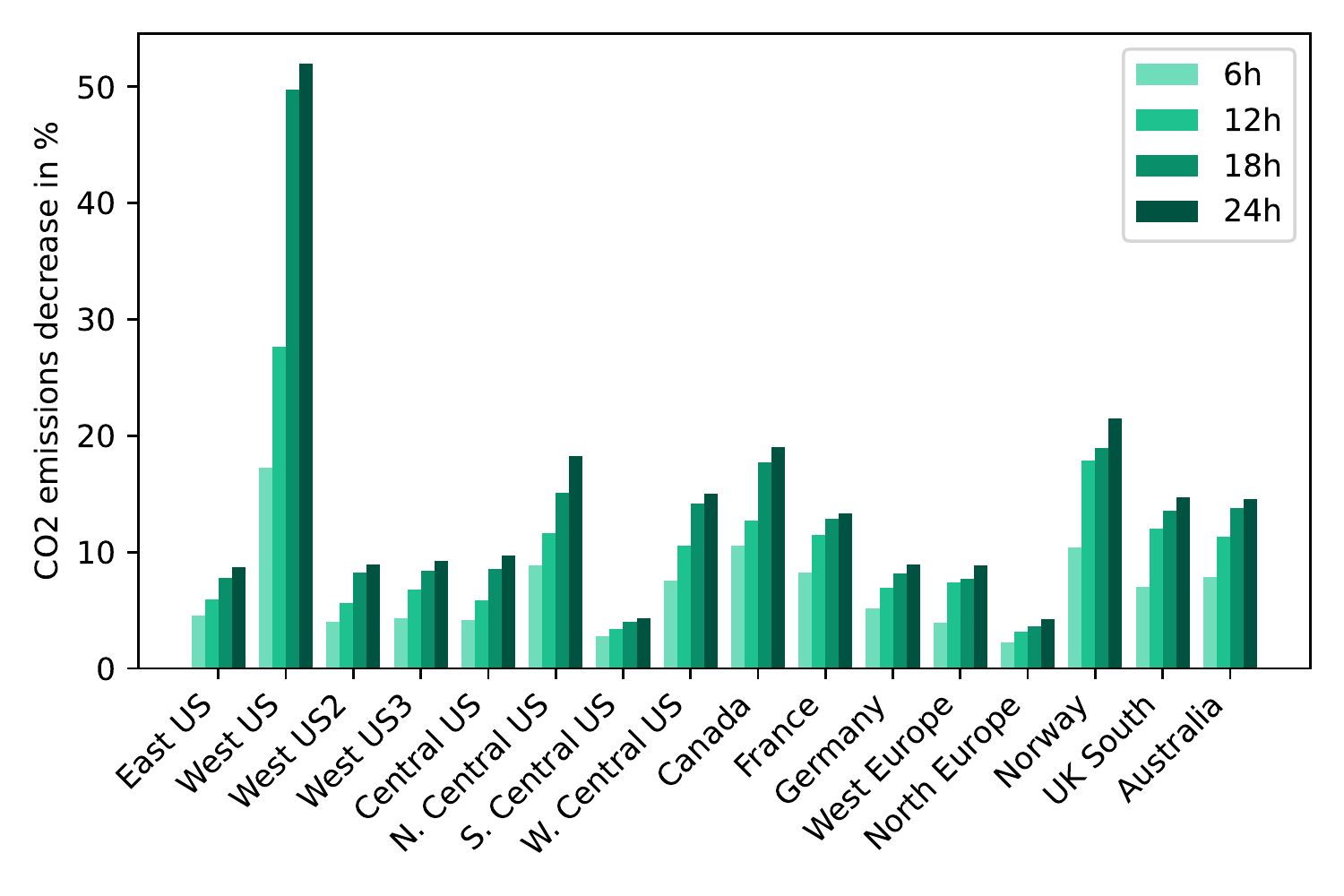}}
    \subfloat[\textit{Pause and Resume} optimization.]{\includegraphics[width=75mm]{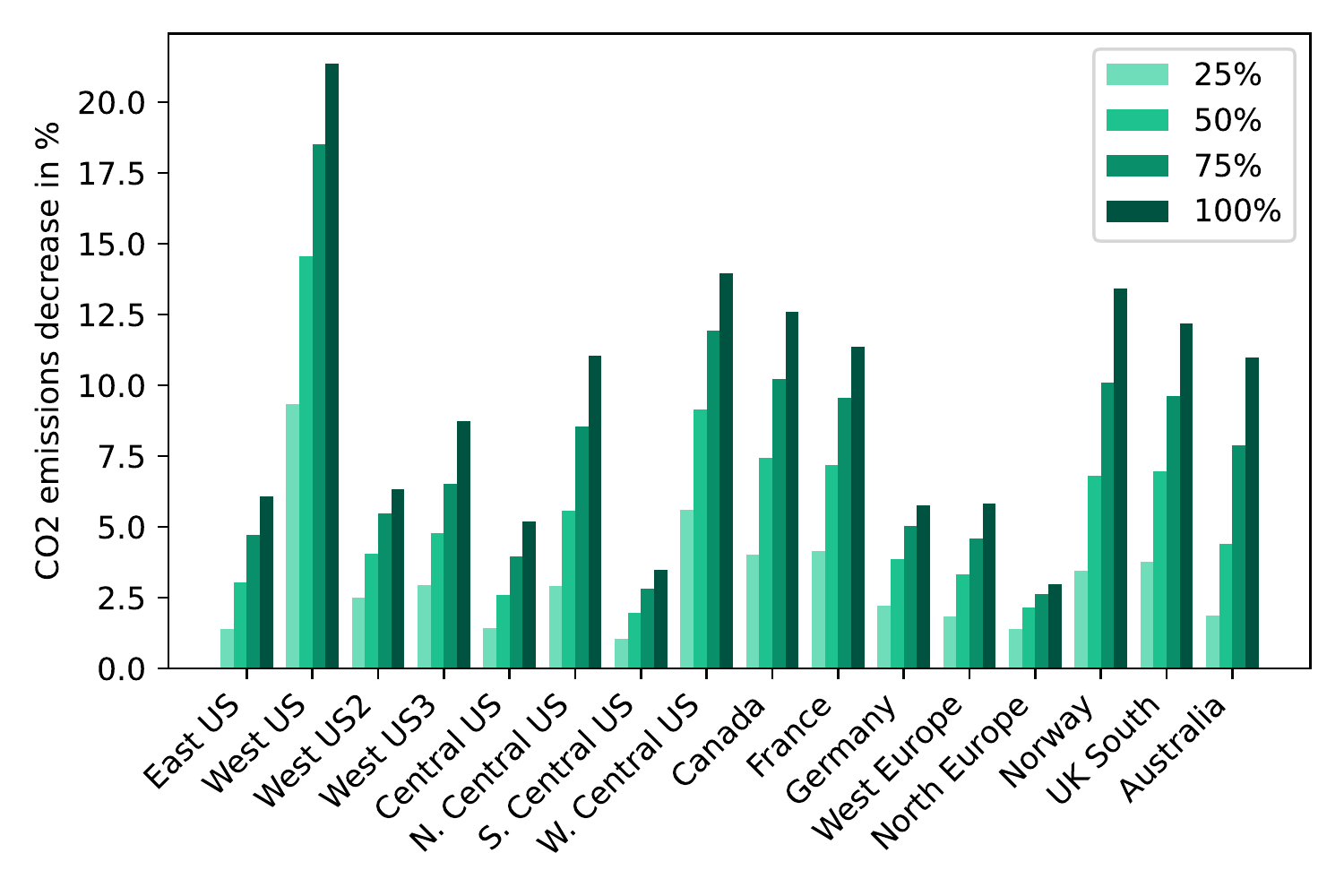}}
    \caption{Optimization results for the finetuning of BERT small on the MNLI dataset, using 4 V100s. Without optimization, the job ran for approximately 6 hours and consumed 3.1 kWh.}
    \label{fig:bert_finetuning}
\end{figure}

\begin{figure}[h]
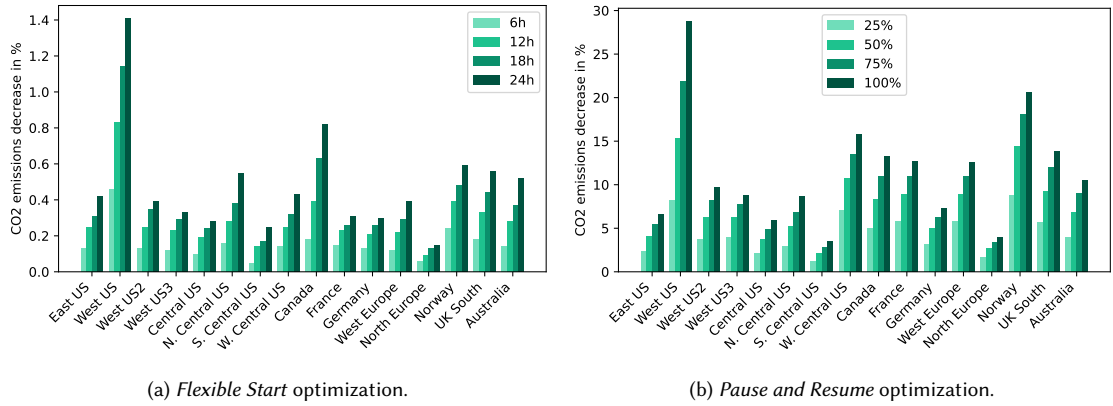

    \centering
    \subfloat[\textit{Flexible Start} optimization.]{\includegraphics[width=75mm]{start_txl.pdf}}
    \subfloat[\textit{Pause and Resume} optimization.]{\includegraphics[width=75mm]{pause_resume_txl.pdf}}
    \caption{Optimization results for the training of a 6B Parameter Transformer on 256 A100s. Without optimization, the job ran for approximately 8 days and consumed 13,812 kWh.}
    \label{fig:txl}
\end{figure}

\begin{figure}[h]
    \centering
    \subfloat[\textit{Flexible Start} optimization.]{\includegraphics[width=75mm]{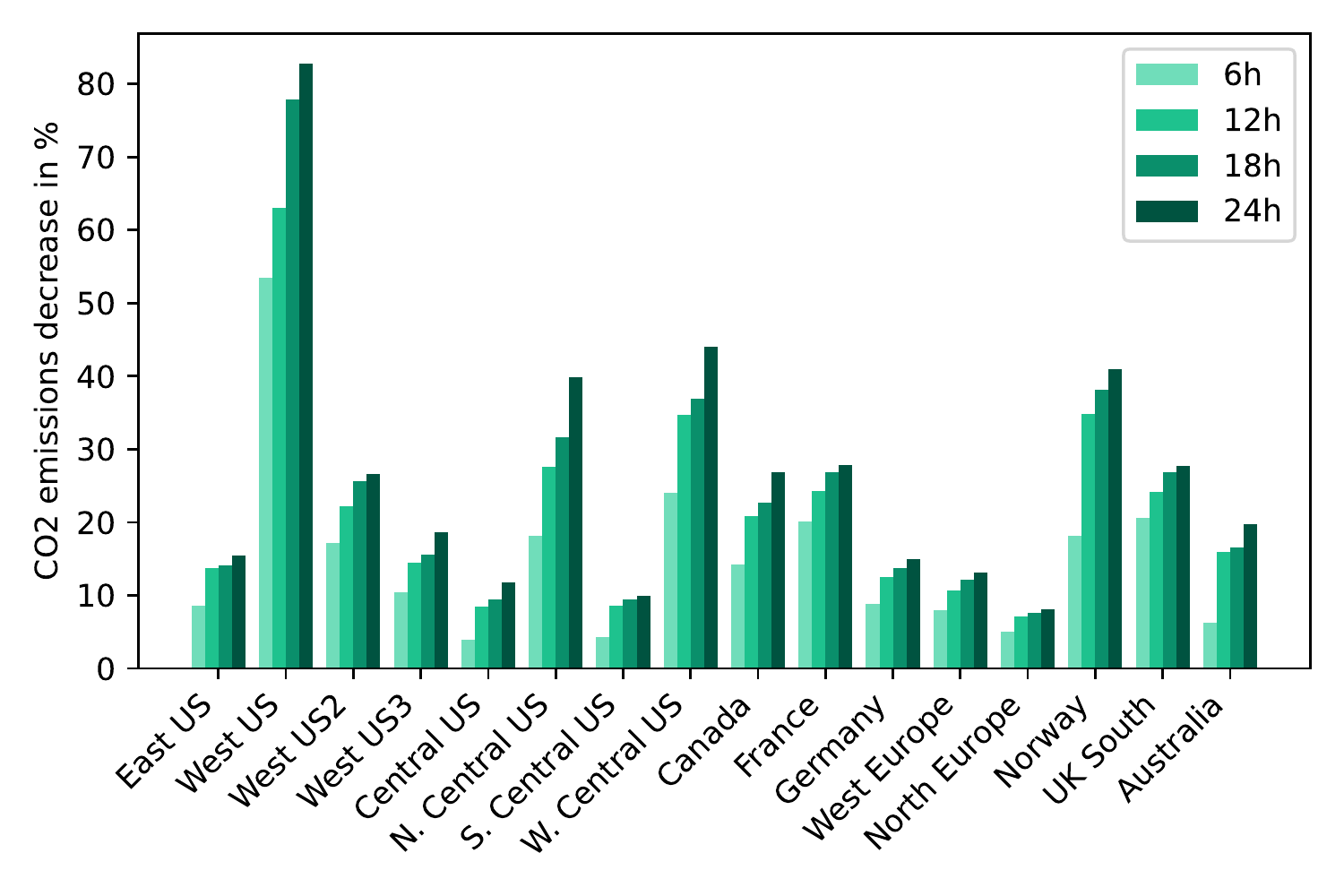}}
    \subfloat[\textit{Pause and Resume} optimization.]{\includegraphics[width=75mm]{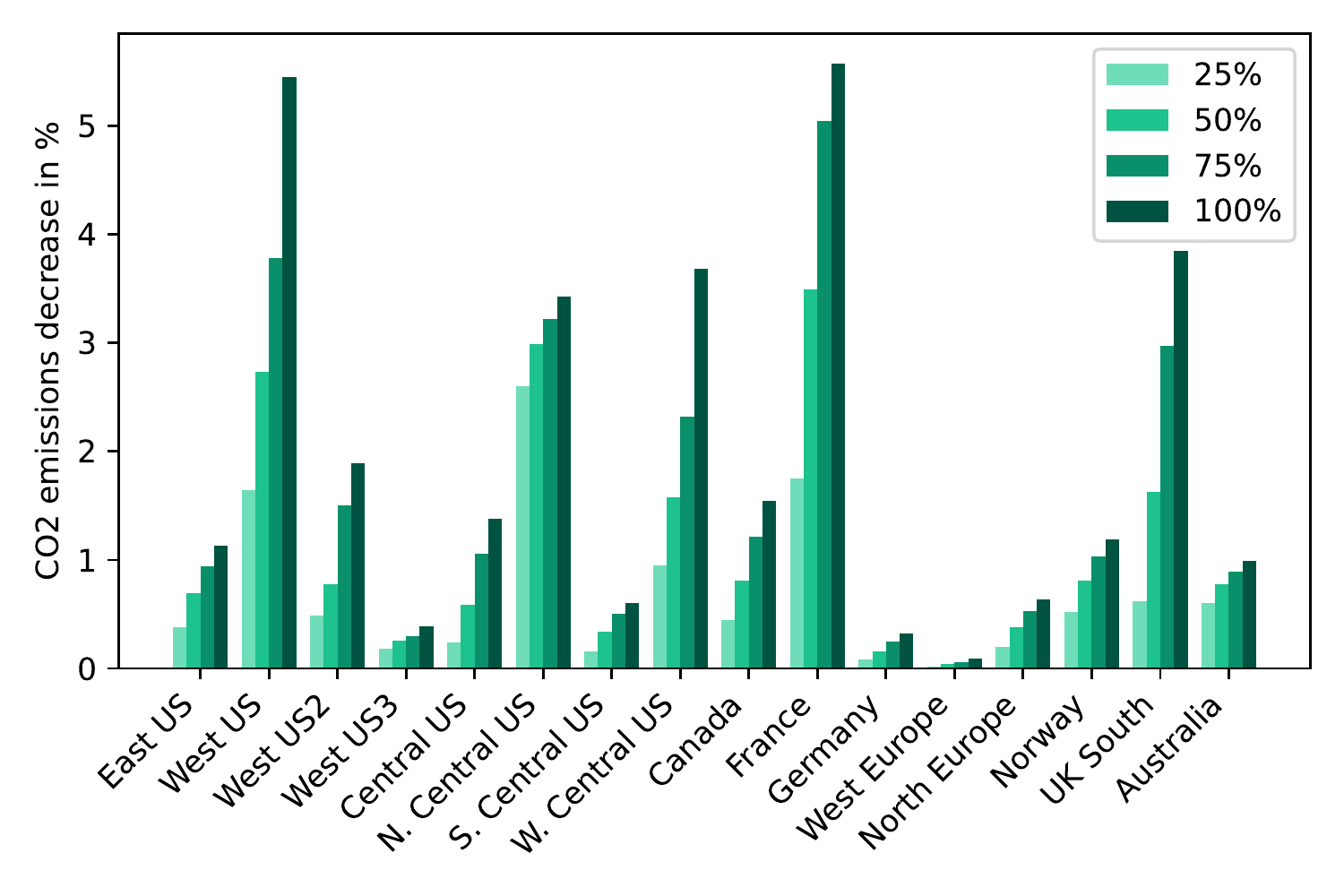}}
    \caption{Optimization results for a DenseNet 121 trained on MNIST on 1 P40. Without optimization, the job ran for approximately 20 minutes and consumed 20 WH.}
    \label{fig:dense_121}
\end{figure}

\begin{figure}[h]
    \centering
    \subfloat[\textit{Flexible Start} optimization.]{\includegraphics[width=75mm]{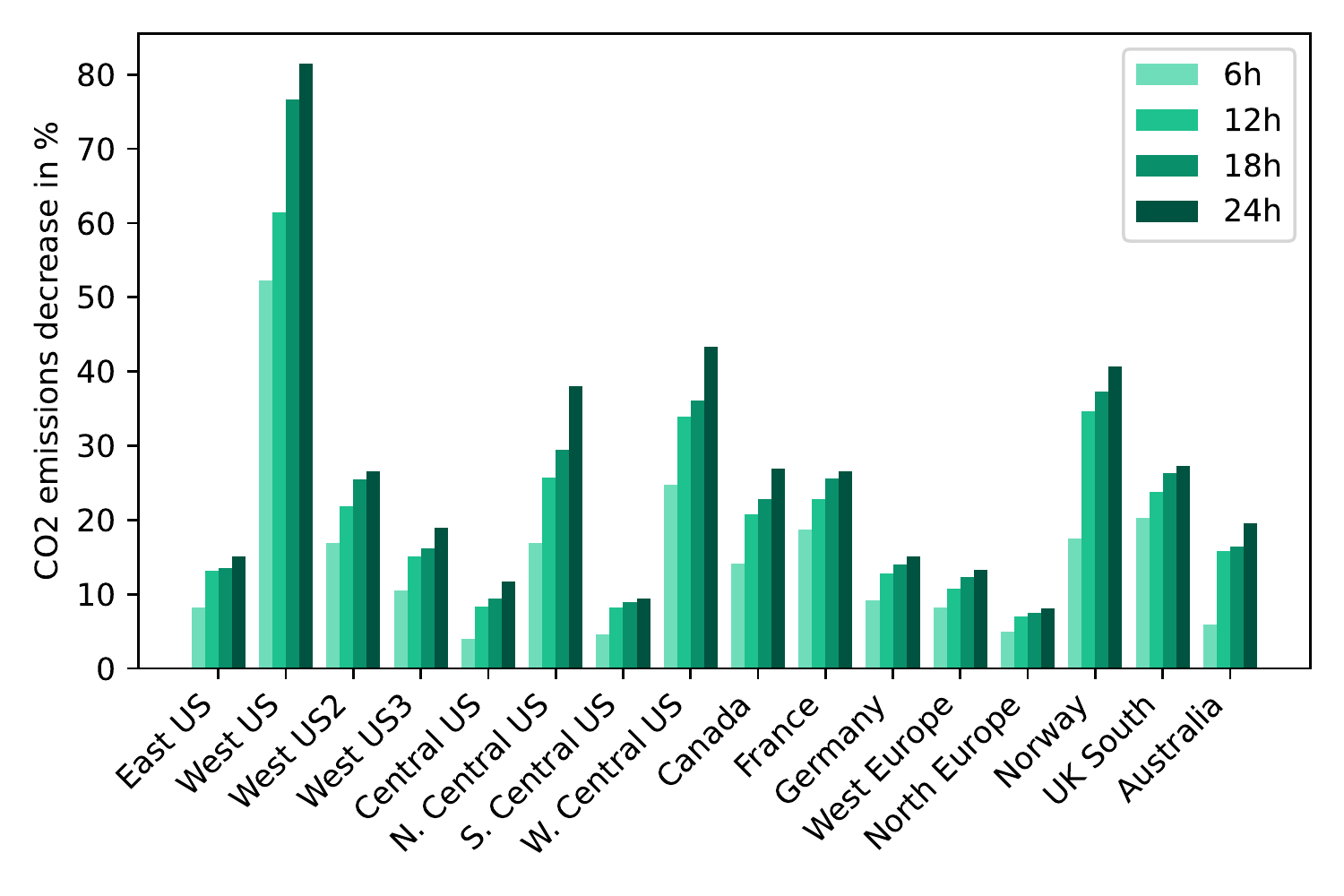}}
    \subfloat[\textit{Pause and Resume} optimization.]{\includegraphics[width=75mm]{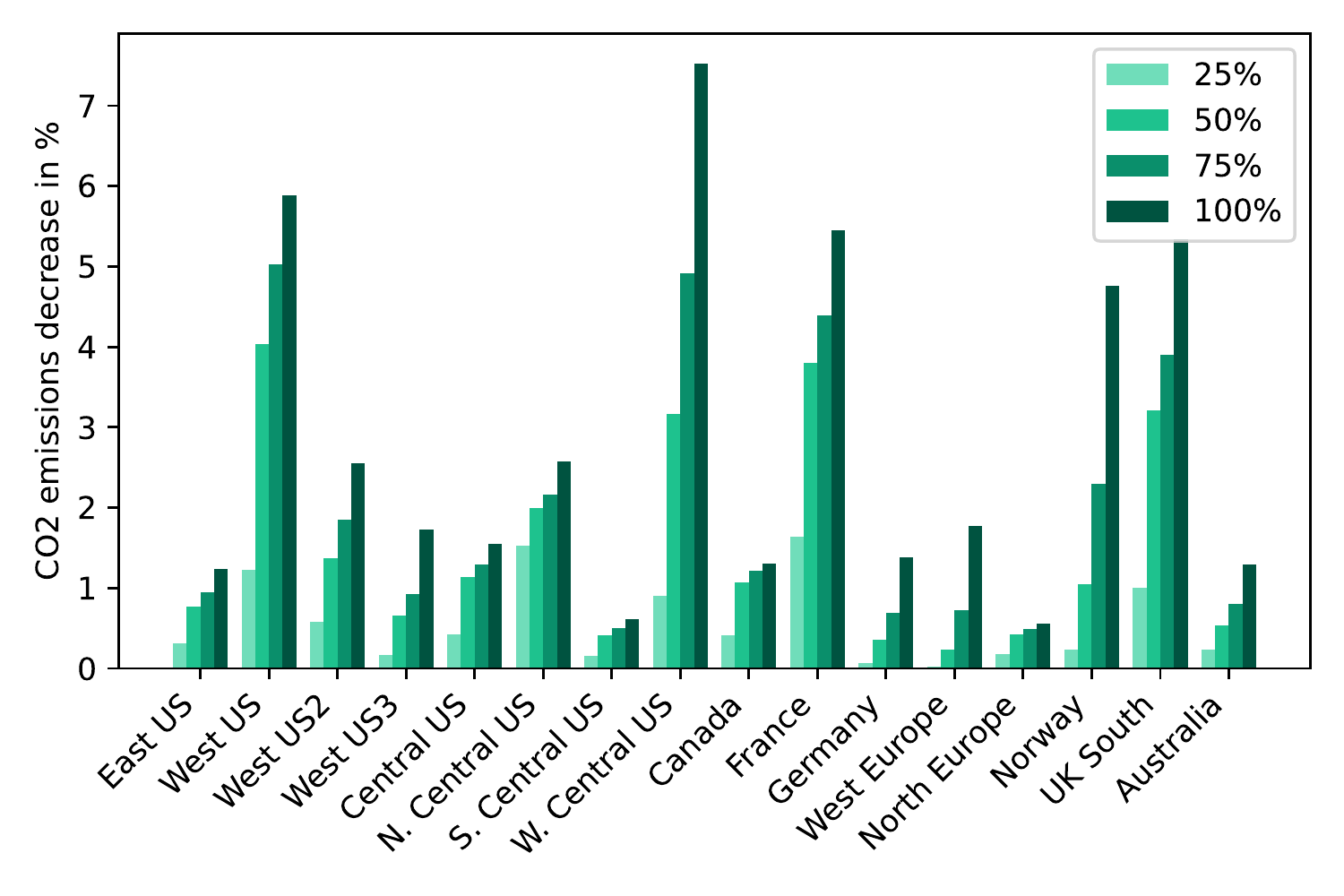}}
    \caption{Optimization results for a DenseNet 169 trained on MNIST on 1 P40. Without optimization, the job ran for approximately 20 minutes and consumed 28 WH.}
    \label{fig:dense_169}
\end{figure}

\begin{figure}[h]
    \centering
    \subfloat[\textit{Flexible Start} optimization.]{\includegraphics[width=75mm]{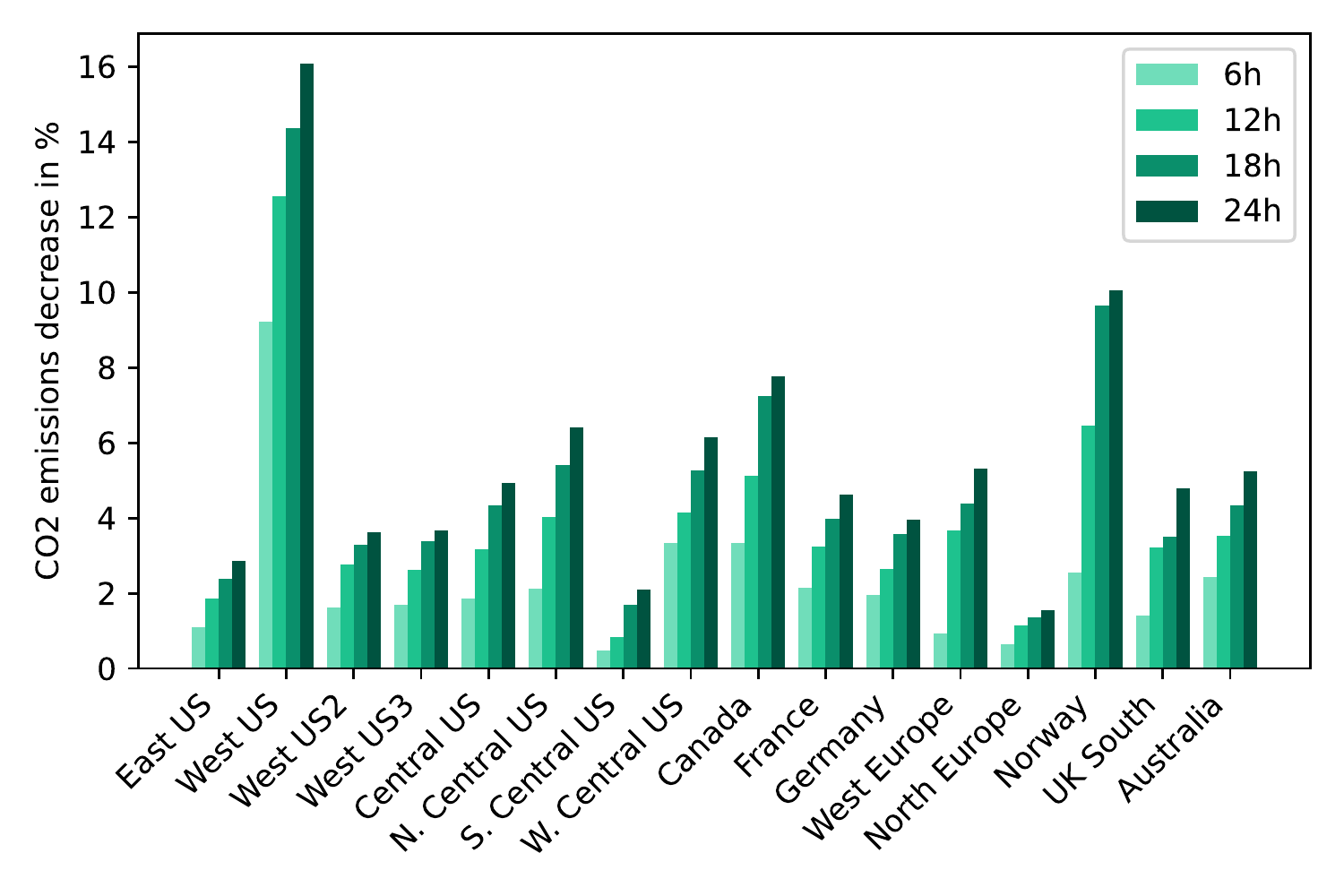}}
    \subfloat[\textit{Pause and Resume} optimization.]{\includegraphics[width=75mm]{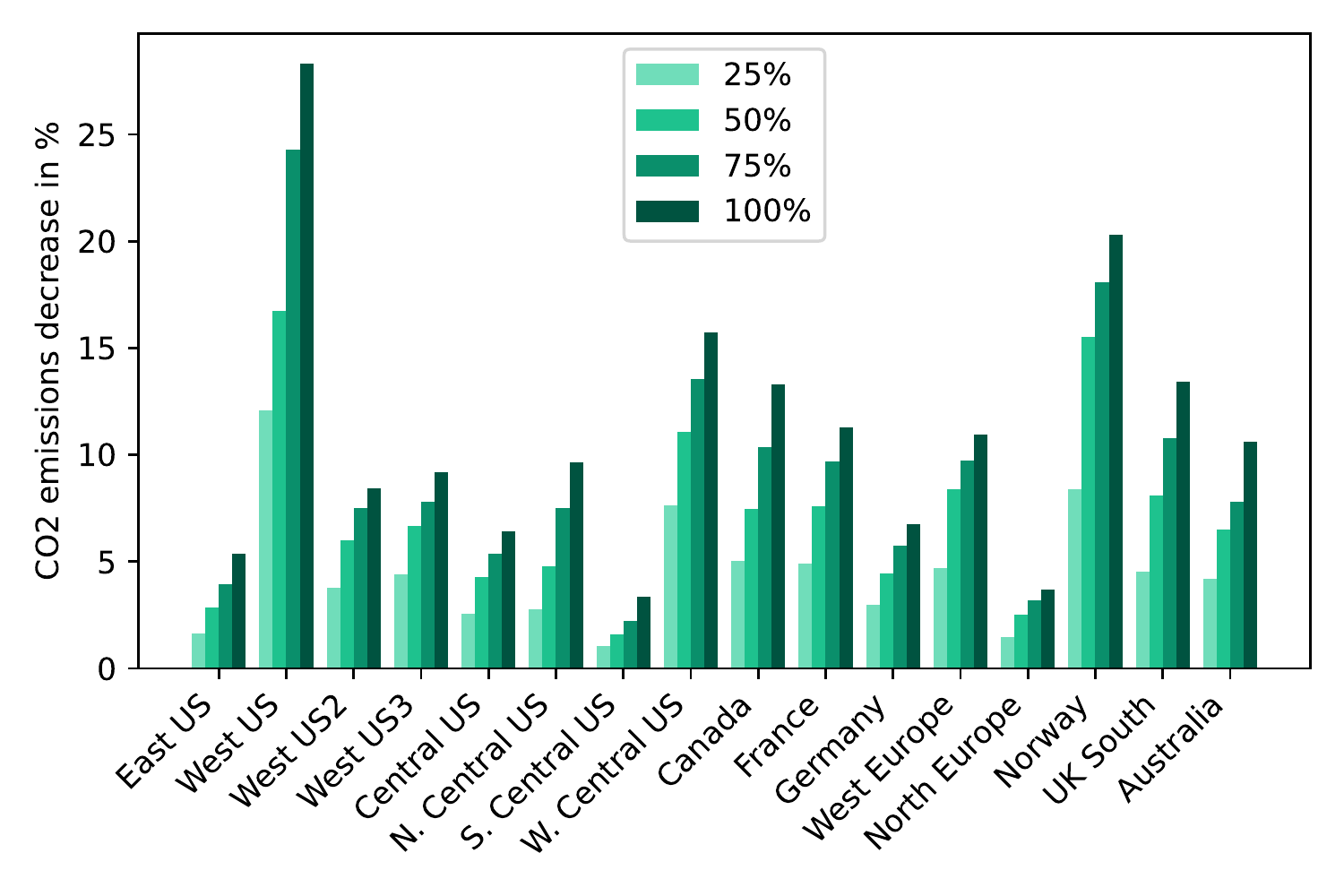}}
    \caption{Optimization results for a Tiny ViT trained on 1 V100. Without optimization, the job ran for approximately 19 hours and consumed 1.7 kWh.}
    \label{fig:vit_tiny}
\end{figure}

\begin{figure}[h]
    \centering
    \subfloat[\textit{Flexible Start} optimization.]{\includegraphics[width=75mm]{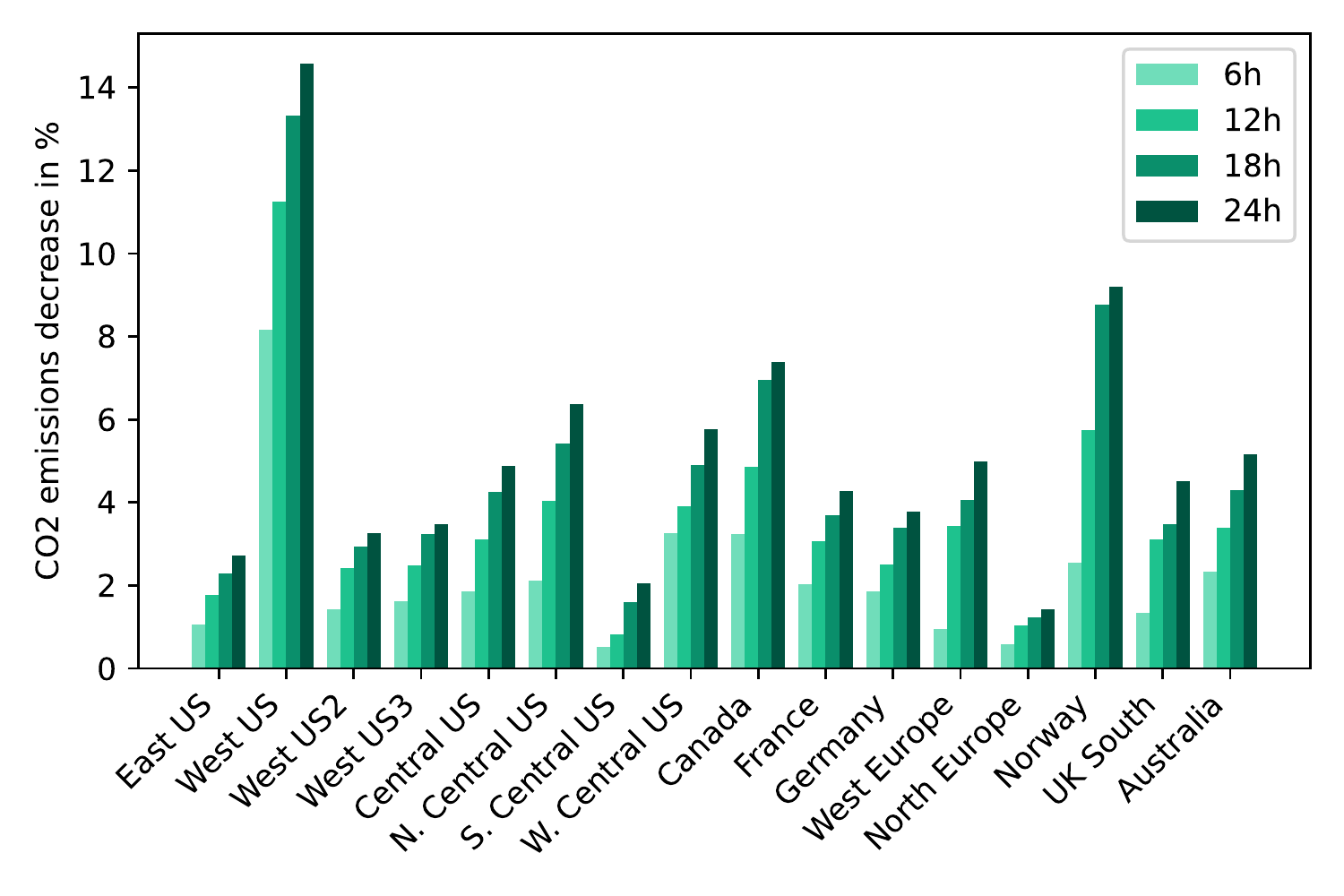}}
    \subfloat[\textit{Pause and Resume} optimization.]{\includegraphics[width=75mm]{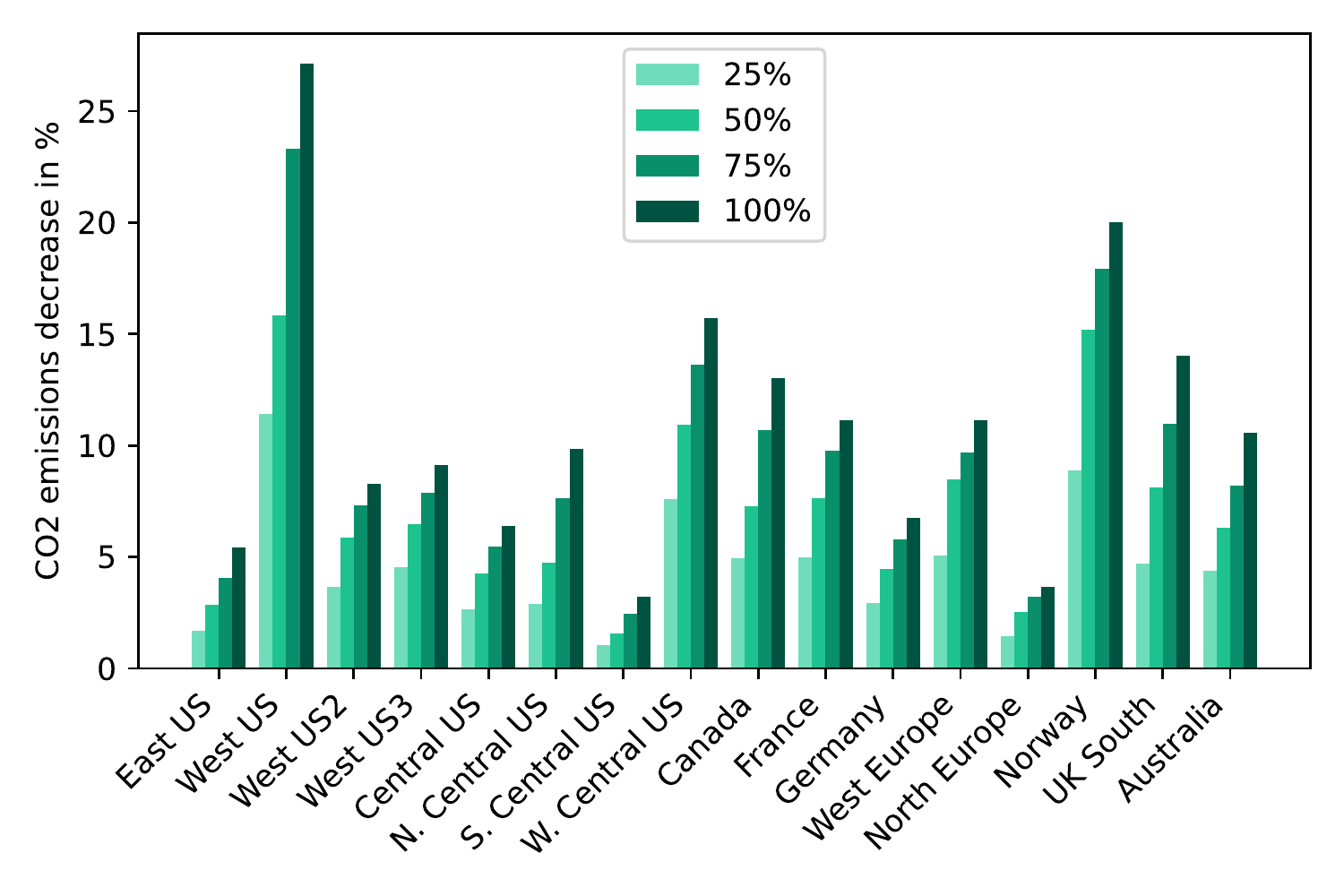}}
    \caption{Optimization results for a Small ViT trained on 1 V100. Without optimization, the job ran for approximately 19 hours and consumed 2.2 kWh.}
    \label{fig:vit_small}
\end{figure}

\begin{figure}[h]
    \centering
    \subfloat[\textit{Flexible Start} optimization.]{\includegraphics[width=75mm]{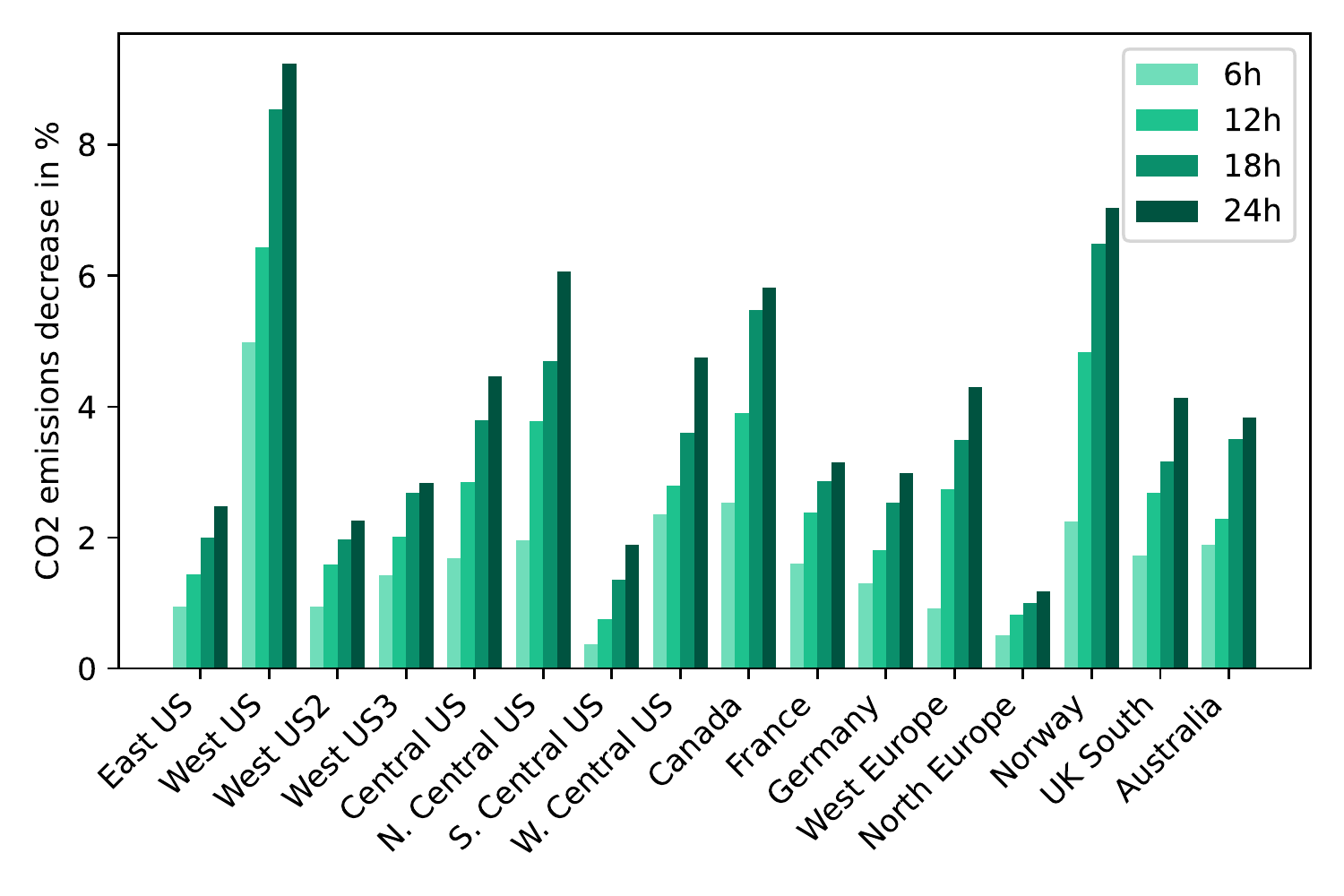}}
    \subfloat[\textit{Pause and Resume} optimization.]{\includegraphics[width=75mm]{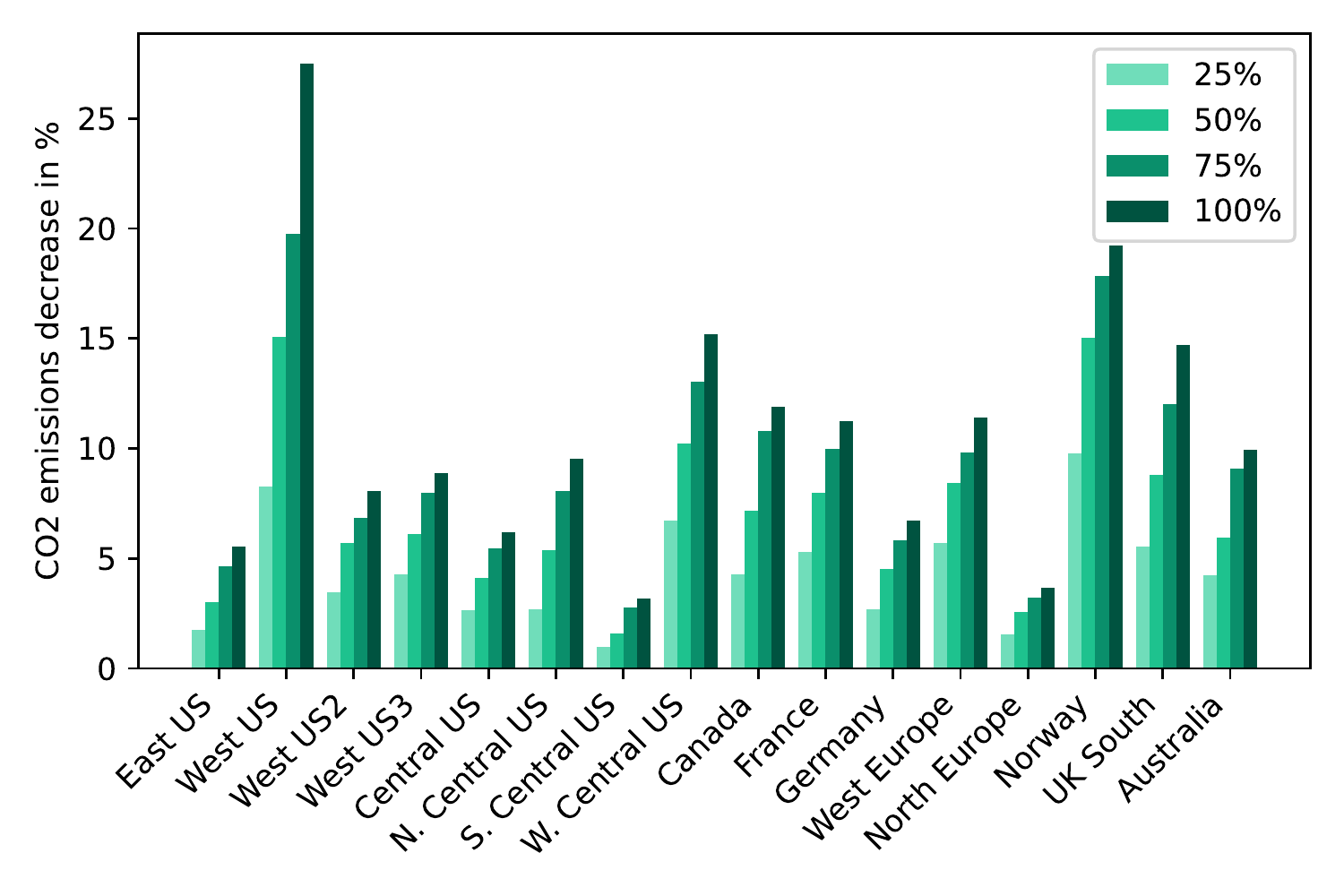}}
    \caption{Optimization results for a Base ViT trained on 1 V100. Without optimization, the job ran for approximately 21 hours and consumed 4.7 kWh.}
    \label{fig:vit_base}
\end{figure}

\begin{figure}[h]
    \centering
    \subfloat[\textit{Flexible Start} optimization.]{\includegraphics[width=75mm]{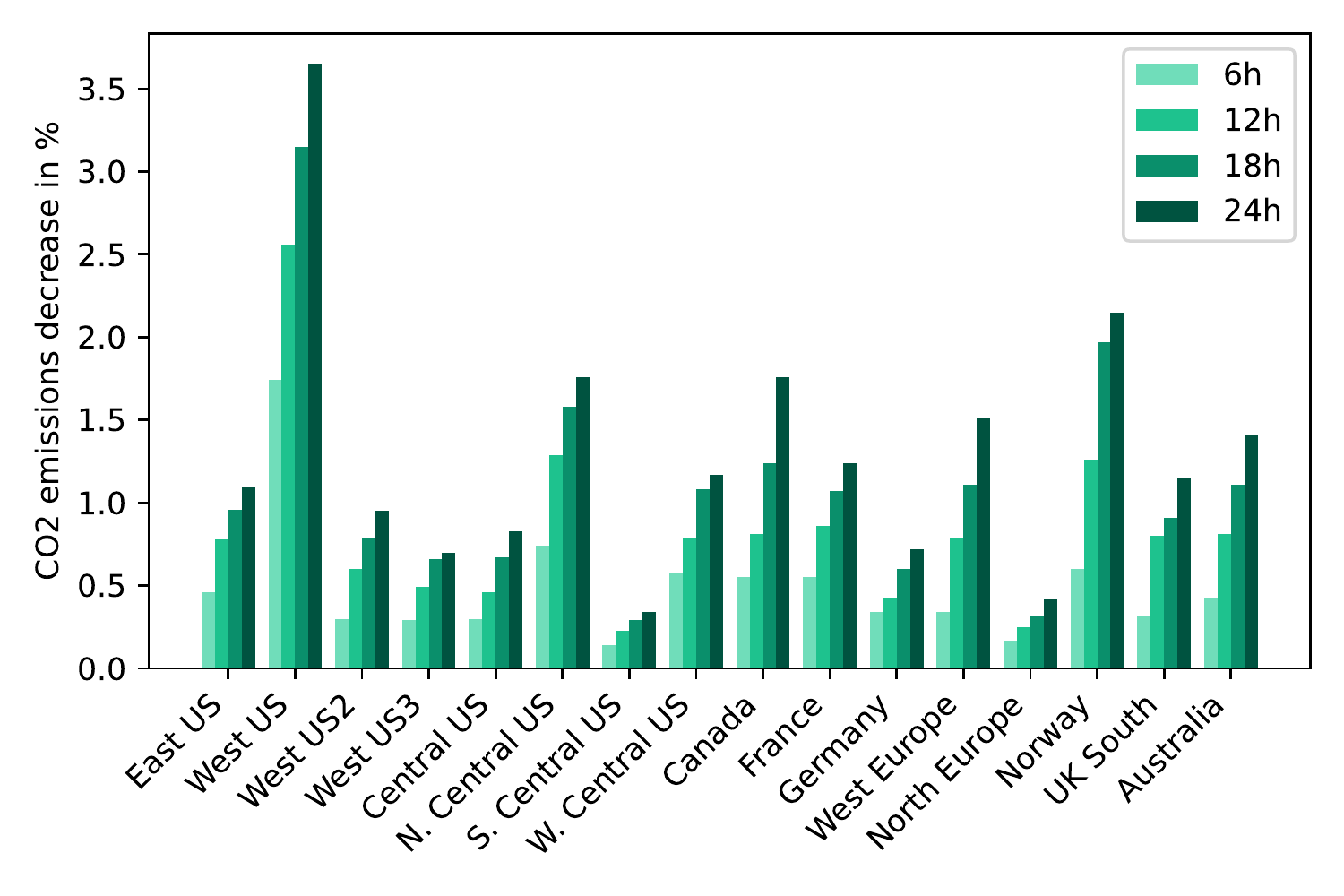}}
    \subfloat[\textit{Pause and Resume} optimization.]{\includegraphics[width=75mm]{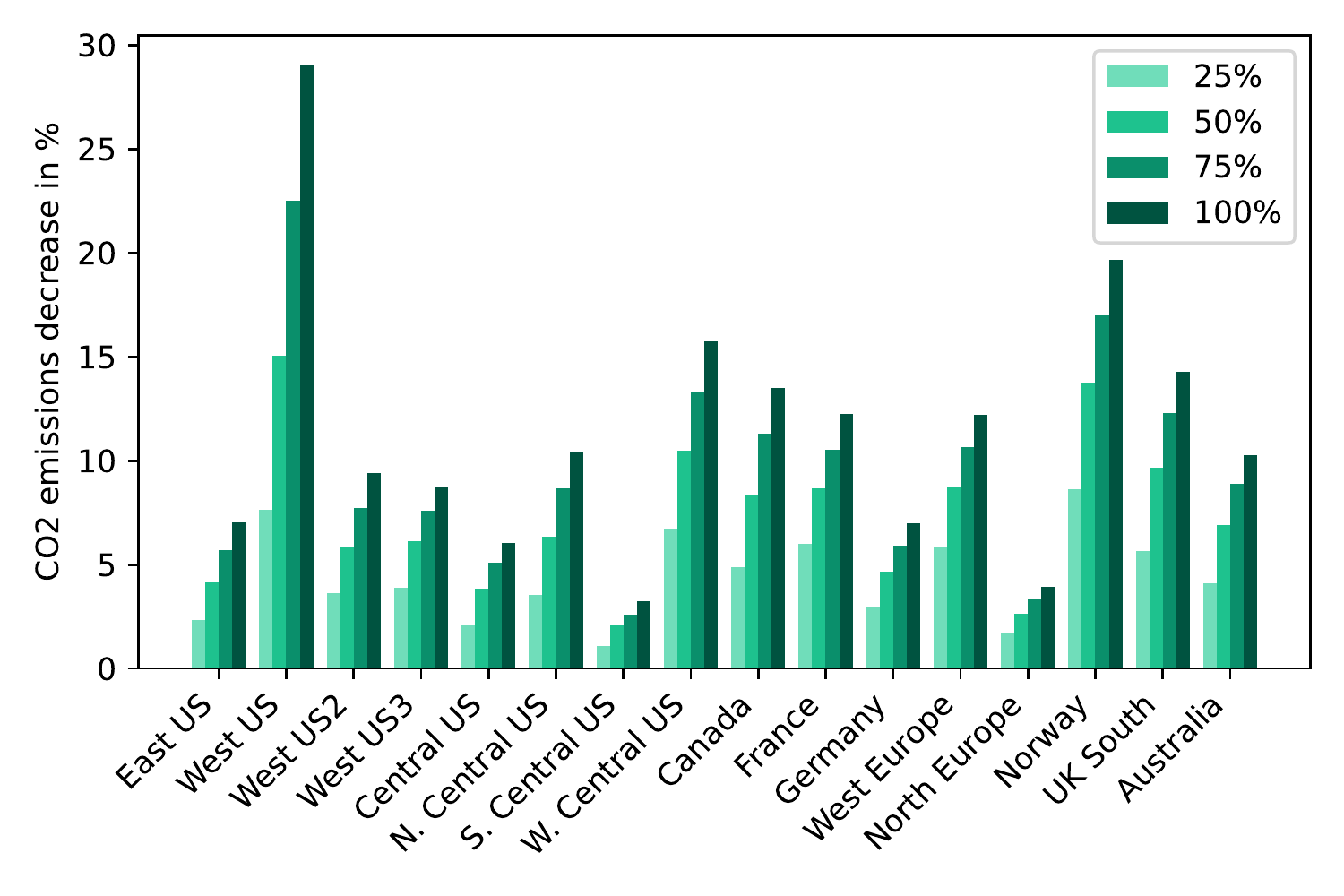}}
    \caption{Optimization results for a Large ViT trained on 4 V100. Without optimization, the job ran for approximately 90 hours and consumed 93.3 kWh.}
    \label{fig:vit_large}
\end{figure}

\begin{figure}[h]
    \centering
    \subfloat[\textit{Flexible Start} optimization.]{\includegraphics[width=75mm]{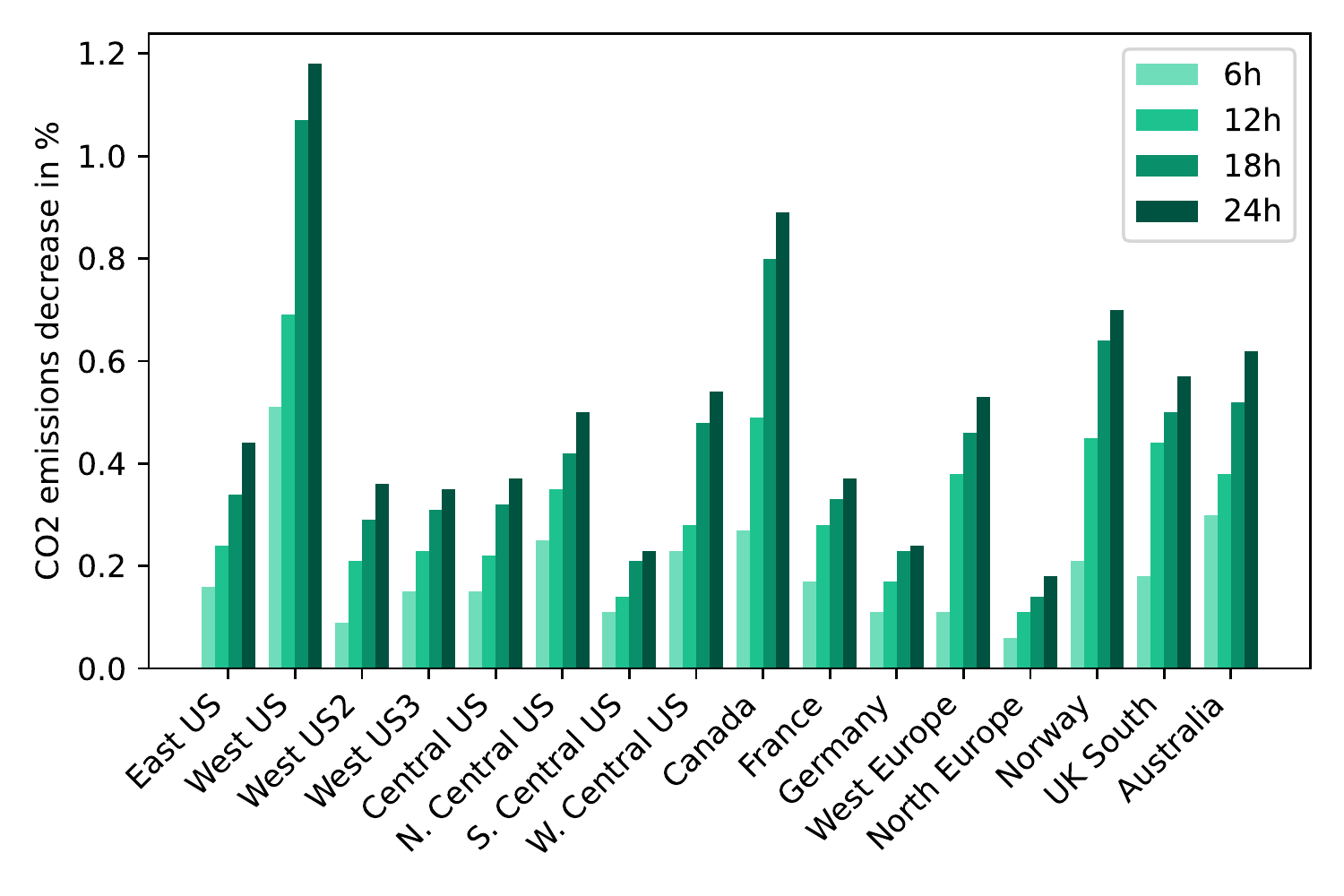}}
    \subfloat[\textit{Pause and Resume} optimization.]{\includegraphics[width=75mm]{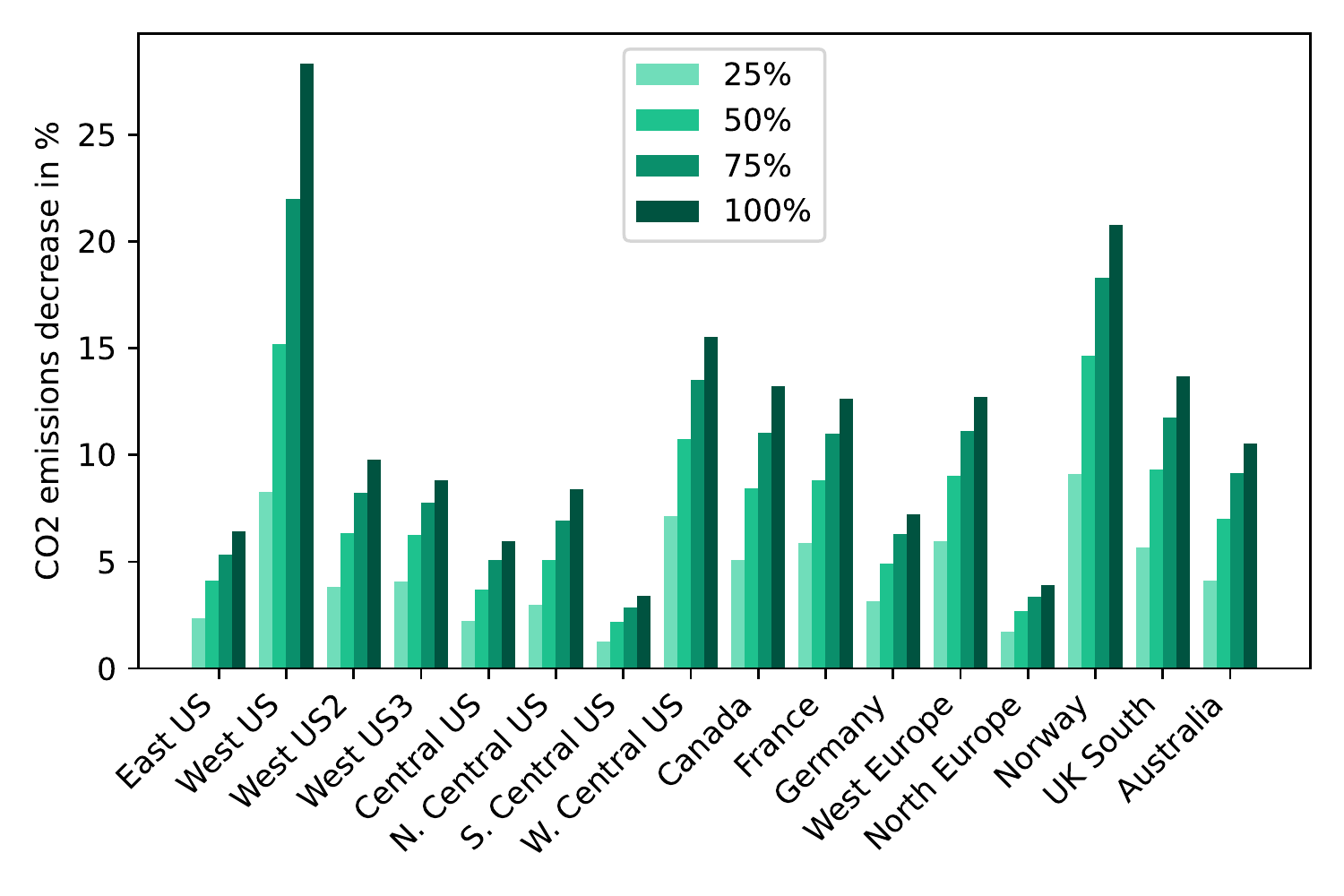}}
    \caption{Optimization results for a Huge ViT trained on 4 V100. Without optimization, the job ran for approximately 9 days and consumed 237.6 kWh.}
    \label{fig:vit_huge}
\end{figure}

In Figures~\ref{fig:bert_training},~\ref{fig:bert_finetuning},~\ref{fig:txl},~\ref{fig:dense_121},~\ref{fig:dense_169},~\ref{fig:dense_201},~\ref{fig:vit_tiny},~\ref{fig:vit_small},~\ref{fig:vit_base},~\ref{fig:vit_large} and \ref{fig:vit_huge}, we report the decrease in \carbondioxide emissions (in percent) obtained when performing the two optimizations introduced in the main text for all 16 regions, all 11 models, averaged over the year and for various values of the $N$ denoting the increase in job duration stemming from the optimization.

\begin{figure}[h]
    \centering
    \subfloat[\textit{Flexible Start} optimization.]{\includegraphics[width=75mm]{start_densenet201.pdf}}
    \subfloat[\textit{Pause and Resume} optimization.]{\includegraphics[width=75mm]{pause_resume_densenet201.pdf}}
    \caption{Optimization results for a DenseNet 201 trained on MNIST on 1 P40. Without optimization, the job ran for approximately 25 minutes and consumed 37 WH.}
    \label{fig:dense_201}
\end{figure}

\clearpage

\section{Additional Tables}

In Tables~\ref{tab:25},~\ref{tab:50},~\ref{tab:75},~\ref{tab:100},~\ref{tab:6},~\ref{tab:12},~\ref{tab:18} and~\ref{tab:24}, we report the decrease in \carbondioxide emissions (in percent) obtained when performing the two optimizations introduced in the main text for all 11 models, averaged across the 16 regions we consider and over the year, for various values of the $N$ denoting the increase in job duration stemming from the optimization.

\begin{table}[h]
\centering
\begin{tabular}{ | c||c|c|c||c|c|c||c|c|c|c|c | } 
\hline
Model & BERT  & BERT  & 6B & Dense  & Dense  & Dense  & ViT  & ViT  & ViT  & ViT  & ViT \\
 & finetune & LM & Transf. & 121 & 169 & 201 & Tiny & Small & Base & Large & Huge \\
\hline
FS & 1.9\% & 1.7\% & 0.8\% & 0.6\% & 0.4\% & 0.4\% & 1.8\% & 1.8\% & 1.6\% & 1.3\% & 0.9\% \\
\hline
P\&R  & 3.1\% & 4.1\% & 4.5\% & 0.7\% & 0.6\% & 0.5\% & 4.5\% & 4.6\% & 4.4\% & 4.4\% & 4.5\% \\
\hline
Pauses  / hr & 0.45 & 0.25 & 0.22 & 2.2 & 2.4 & 2.4 & 0.28 & 0.29 & 0.28 & 0.22 & 0.21 \\
\hline
\end{tabular}
\caption{For the 11 models in our analysis: the gain in percent averaged over the year and across the 16 regions for the \textit{Flexible Start} (FS) and \textit{Pause and Resume} (P\&R) optimizations allowing for a 25\% increase in job duration. The last line represents the average number of pauses per hour performed by the P\&R optimization.}
\label{tab:25}
\end{table}

\begin{table}[h]
\centering
\begin{tabular}{ | c||c|c|c||c|c|c||c|c|c|c|c | } 
\hline
Model & BERT  & BERT  & 6B & Dense  & Dense  & Dense  & ViT  & ViT  & ViT  & ViT  & ViT \\
 & finetune & LM & Transf. & 121 & 169 & 201 & Tiny & Small & Base & Large & Huge \\
\hline
FS & 3.6\% & 2.9\% & 1.5\% & 1.0\% & 1.3\% & 1.2\% & 3.3\% & 3.1\% & 2.5\% & 2.1\% & 1.6\% \\
\hline
P\&R  & 5.5\% & 7.0\% & 7.4\% & 1.1\% & 1.5\% & 1.6\% & 7.2\% & 7.0\% & 7.0\% & 7.3\% & 7.4\% \\
\hline
Pauses  / hr & 0.47 & 0.29 & 0.27 & 1.83 & 2.33 & 2.33 & 0.32 & 0.33 & 0.32 & 0.27 & 0.26 \\
\hline
\end{tabular}
\caption{For the 11 models in our analysis: the gain in percent averaged over the year and across the 16 regions for the \textit{Flexible Start} (FS) and \textit{Pause and Resume} (P\&R) optimizations allowing for a 50\% increase in job duration. The last line represents the average number of pauses per hour performed by the P\&R optimization.}
\label{tab:50}
\end{table}

\begin{table}[h]
\centering
\begin{tabular}{ | c||c|c|c||c|c|c||c|c|c|c|c | } 
\hline
Model & BERT  & BERT  & 6B & Dense  & Dense  & Dense  & ViT  & ViT  & ViT  & ViT  & ViT \\
 & finetune & LM & Transf. & 121 & 169 & 201 & Tiny & Small & Base & Large & Huge \\
\hline
FS & 5.4\% & 3.6\% & 2.0\% & 1.5\% & 1.7\% & 1.9\% & 4.2\% & 4.0\% & 3.3\% & 2.8\% & 2.2\% \\
\hline
P\&R  & 7.6\% & 9.2\% & 9.6\% & 1.6\% & 2.0\% & 2.4\% & 9.2\% & 9.3\% & 9.2\% & 9.6\% & 9.6\% \\
\hline
Pauses  / hr & 0.45 & 0.3 & 0.27 & 1.71 & 2.0 & 2.14 & 0.33 & 0.33 & 0.32 & 0.28 & 0.26 \\
\hline
\end{tabular}
\caption{For the 11 models in our analysis: the gain in percent averaged over the year and across the 16 regions for the \textit{Flexible Start} (FS) and \textit{Pause and Resume} (P\&R) optimizations allowing for a 75\% increase in job duration. The last line represents the average number of pauses per hour performed by the P\&R optimization.}
\label{tab:75}
\end{table}

\begin{table}[h]
\centering
\begin{tabular}{ | c||c|c|c||c|c|c||c|c|c|c|c | } 
\hline
Model & BERT  & BERT  & 6B & Dense  & Dense  & Dense  & ViT  & ViT  & ViT  & ViT  & ViT \\
 & finetune & LM & Transf. & 121 & 169 & 201 & Tiny & Small & Base & Large & Huge \\
\hline
FS & 7.0\% & 4.1\% & 2.6\% & 1.8\% & 2.5\% & 2.7\% & 5.0\% & 4.8\% & 3.9\% & 3.3\% & 3.0\% \\
\hline
P\&R  & 9.5\% & 11.0\% & 11.4\% & 2.0\% & 2.8\% & 3.1\% & 11.0\% & 11.0\% & 10.8\% & 11.4\% & 11.3\% \\
\hline
Pauses  / hr & 0.42 & 0.29 & 0.27 & 1.5 & 1.88 & 2.0 & 0.31 & 0.32 & 0.31 & 0.27 & 0.26 \\
\hline
\end{tabular}
\caption{For the 11 models in our analysis: the gain in percent averaged over the year and across the 16 regions for the \textit{Flexible Start} (FS) and \textit{Pause and Resume} (P\&R) optimizations allowing for a 100\% increase in job duration. The last line represents the average number of pauses per hour performed by the P\&R optimization.}
\label{tab:100}
\end{table}

\begin{table}[h]
\centering
\begin{tabular}{ | c||c|c|c||c|c|c||c|c|c|c|c | } 
\hline
Model & BERT  & BERT  & 6B & Dense  & Dense  & Dense  & ViT  & ViT  & ViT  & ViT  & ViT \\
 & finetune & LM & Transf. & 121 & 169 & 201 & Tiny & Small & Base & Large & Huge \\
\hline
FS & 6.9\% & 1.2\% & 0.2\% & 15.3\% & 14.9\% & 14.5\% & 2.3\% & 2.2\% & 1.7\% & 0.5\% & 0.2\% \\
\hline
P\&R  & 9.4\% & 2.9\% & 0.8\% & 15.8\% & 15.5\% & 15.3\% & 5.5\% & 5.3\% & 4.8\% & 1.5\% & 0.7\% \\
\hline
Pauses  / hr & 0.41 & 0.21 & 0.06 & 0.22 & 0.27 & 0.28 & 0.29 & 0.3 & 0.29 & 0.11 & 0.06 \\
\hline
\end{tabular}
\caption{For the 11 models in our analysis: the gain in percent averaged over the year and across the 16 regions for the \textit{Flexible Start} (FS) and \textit{Pause and Resume} (P\&R) optimizations allowing for a 6h increase in job duration. The last line represents the average number of pauses per hour performed by the P\&R optimization.}
\label{tab:6}
\end{table}

\begin{table}[h]
\centering
\begin{tabular}{ | c||c|c|c||c|c|c||c|c|c|c|c | } 
\hline
Model & BERT  & BERT  & 6B & Dense  & Dense  & Dense  & ViT  & ViT  & ViT  & ViT  & ViT \\
 & finetune & LM & Transf. & 121 & 169 & 201 & Tiny & Small & Base & Large & Huge \\
\hline
FS & 10.1\% & 2.3\% & 0.3\% & 21.6\% & 21.1\% & 20.6\% & 3.8\% & 3.6\% & 2.7\% & 0.8\% & 0.3\% \\
\hline
P\&R  & 13.8\% & 5.3\% & 1.4\% & 22.2\% & 21.7\% & 21.5\% & 8.3\% & 8.1\% & 7.7\% & 2.7\% & 1.3\% \\
\hline
Pauses  / hr & 0.33 & 0.27 & 0.1 & 0.12 & 0.15 & 0.15 & 0.32 & 0.33 & 0.32 & 0.17 & 0.09 \\
\hline
\end{tabular}
\caption{For the 11 models in our analysis: the gain in percent averaged over the year and across the 16 regions for the \textit{Flexible Start} (FS) and \textit{Pause and Resume} (P\&R) optimizations allowing for a 12h increase in job duration. The last line represents the average number of pauses per hour performed by the P\&R optimization.}
\label{tab:12}
\end{table}

\begin{table}[h]
\centering
\begin{tabular}{ | c||c|c|c||c|c|c||c|c|c|c|c | } 
\hline
Model & BERT  & BERT  & 6B & Dense  & Dense  & Dense  & ViT  & ViT  & ViT  & ViT  & ViT \\
 & finetune & LM & Transf. & 121 & 169 & 201 & Tiny & Small & Base & Large & Huge \\
\hline
FS & 13.3\% & 2.9\% & 0.4\% & 24.1\% & 23.7\% & 23.2\% & 4.9\% & 4.6\% & 3.6\% & 1.1\% & 0.4\% \\
\hline
P\&R  & 17.4\% & 7.0\% & 2.0\% & 24.9\% & 24.5\% & 24.2\% & 10.8\% & 10.5\% & 9.9\% & 3.8\% & 1.9\% \\
\hline
Pauses  / hr & 0.26 & 0.29 & 0.13 & 0.08 & 0.09 & 0.1 & 0.31 & 0.32 & 0.32 & 0.2 & 0.12 \\
\hline
\end{tabular}
\caption{For the 11 models in our analysis: the gain in percent averaged over the year and across the 16 regions for the \textit{Flexible Start} (FS) and \textit{Pause and Resume} (P\&R) optimizations allowing for a 18h increase in job duration. The last line represents the average number of pauses per hour performed by the P\&R optimization.}
\label{tab:18}
\end{table}

\begin{table}[h]
\centering
\begin{tabular}{ | c||c|c|c||c|c|c||c|c|c|c|c | } 
\hline
Model & BERT  & BERT  & 6B & Dense  & Dense  & Dense  & ViT  & ViT  & ViT  & ViT  & ViT \\
 & finetune & LM & Transf. & 121 & 169 & 201 & Tiny & Small & Base & Large & Huge \\
\hline
FS & 14.5\% & 3.4\% & 0.5\% & 26.8\% & 26.4\% & 25.9\% & 5.6\% & 5.3\% & 4.2\% & 1.3\% & 0.5\% \\
\hline
P\&R  & 19.0\% & 8.5\% & 2.5\% & 27.7\% & 27.3\% & 27.1\% & 12.5\% & 12.3\% & 11.7\% & 4.7\% & 2.4\% \\
\hline
Pauses  / hr & 0.23 & 0.3 & 0.15 & 0.06 & 0.07 & 0.08 & 0.3 & 0.3 & 0.3 & 0.23 & 0.14 \\
\hline
\end{tabular}
\caption{For the 11 models in our analysis: the gain in percent averaged over the year and across the 16 regions for the \textit{Flexible Start} (FS) and \textit{Pause and Resume} (P\&R) optimizations allowing for a 24h increase in job duration. The last line represents the average number of pauses per hour performed by the P\&R optimization.}
\label{tab:24}
\end{table}

\end{document}